\documentclass[10pt,journal,compsoc]{IEEEtran}
\usepackage{epsfig}
\usepackage{graphicx}
\usepackage{amsmath}
\usepackage{amssymb}
\usepackage{bm}
\usepackage{booktabs}
\newcommand{\RNum}[1]{\uppercase\expandafter{\romannumeral #1\relax}}
\usepackage{makecell}
\ifCLASSOPTIONcompsoc

  \usepackage[nocompress]{cite}
\else

  \usepackage{cite}
\fi

\ifCLASSINFOpdf
\else

\fi

\begin{document}

\title{Hierarchical Feature-Aware Tracking}

\author{Wenhua~Zhang,
        Licheng~Jiao,~\IEEEmembership{Fellow,~IEEE,} Fang~Liu, ~\IEEEmembership{Senior Member,~IEEE,}\\Shuyuan~Yang, ~\IEEEmembership{Senior Member,~IEEE},~and~Biao~Hou, ~\IEEEmembership{Member,~IEEE}

\IEEEcompsocitemizethanks{\IEEEcompsocthanksitem This work was supported by the Project supported the Foundation for Innovative Research Groups of the National Natural Science Foundation of China (Grant no. 61621005), the National Natural Science Foundation of China (Grant no. U1701267, 61573267, and 61473215), the Fund for Foreign Scholars in University Research and Teaching Programs the 111 Project©(Grant no. B07048), the Major Research Plan of the National Natural Science Foundation of China (Grant no. 91438201 and 91438103), and the Program for Cheung Kong Scholars and Innovative Research Team in University (Grant no. IRT\_15R53).
\IEEEcompsocthanksitem The authors are with Key Laboratory of Intelligent Perception and Image Understanding of Ministry of Education, International Research Center for Intelligent Perception and Computation, Joint International Research Laboratory of Intelligent Perception and Computation, School of Artificial Intelligence, Xidian University, Xi'an, Shaanxi Province 710071, China.}
}

\markboth{IEEE}%
{Shell \MakeLowercase{\textit{et al.}}: Bare Demo of IEEEtran.cls for Computer Society Journals}

\IEEEtitleabstractindextext{%
\begin{abstract}
In this paper, we propose a hierarchical feature-aware tracking framework for efficient visual tracking. Recent years, ensembled trackers which combine multiple component trackers have achieved impressive performance. In ensembled trackers, the decision of results is usually a post-event process, i.e., tracking result for each tracker is first obtained and then the suitable one is selected according to result ensemble. In this paper, we propose a pre-event method. We construct an expert pool with each expert being one set of features. For each frame, several experts are first selected in the pool according to their past performance and then they are used to predict the object. The selection rate of each expert in the pool is then updated and tracking result is obtained according to result ensemble. We propose a novel pre-known expert-adaptive selection strategy. Since the process is more efficient, more experts can be constructed by fusing more types of features which leads to more robustness. Moreover, with the novel expert selection strategy, overfitting caused by fixed experts for each frame can be mitigated. Experiments on several public available datasets demonstrate the superiority of the proposed method and its state-of-the-art performance among ensembled trackers.
\end{abstract}

\begin{IEEEkeywords}
Hierarchical Feature-Aware Tracking, Ensembled Trackers, Expert Selection Strategy.
\end{IEEEkeywords}}

\maketitle

\IEEEdisplaynontitleabstractindextext

\IEEEpeerreviewmaketitle

\IEEEraisesectionheading{\section{Introduction}\label{sec:introduction}}
\IEEEPARstart{T}{arget} tracking is a fundamental problem in numerous practical computer vision applications \cite{CVPR2017stapleca, my1}. Many breakthroughs have been made on the various tracking datasets, such as OTB-2013 \cite{2013OTB50}, OTB-2015 \cite{2015OTB100}, TempeColor \cite{TC}, UAV123 \cite{UAV123}  and VOT \cite{ECCVW2016VOT, VOT2017, VOT2018}. Despite substantial progress, target tracking remains a challenging problem due to the constantly changing target as well as disturbing scenarios, such as background clutter and occlusion emerges.
\begin{figure}[h]
\begin{center}
\includegraphics[width=1.0\linewidth]{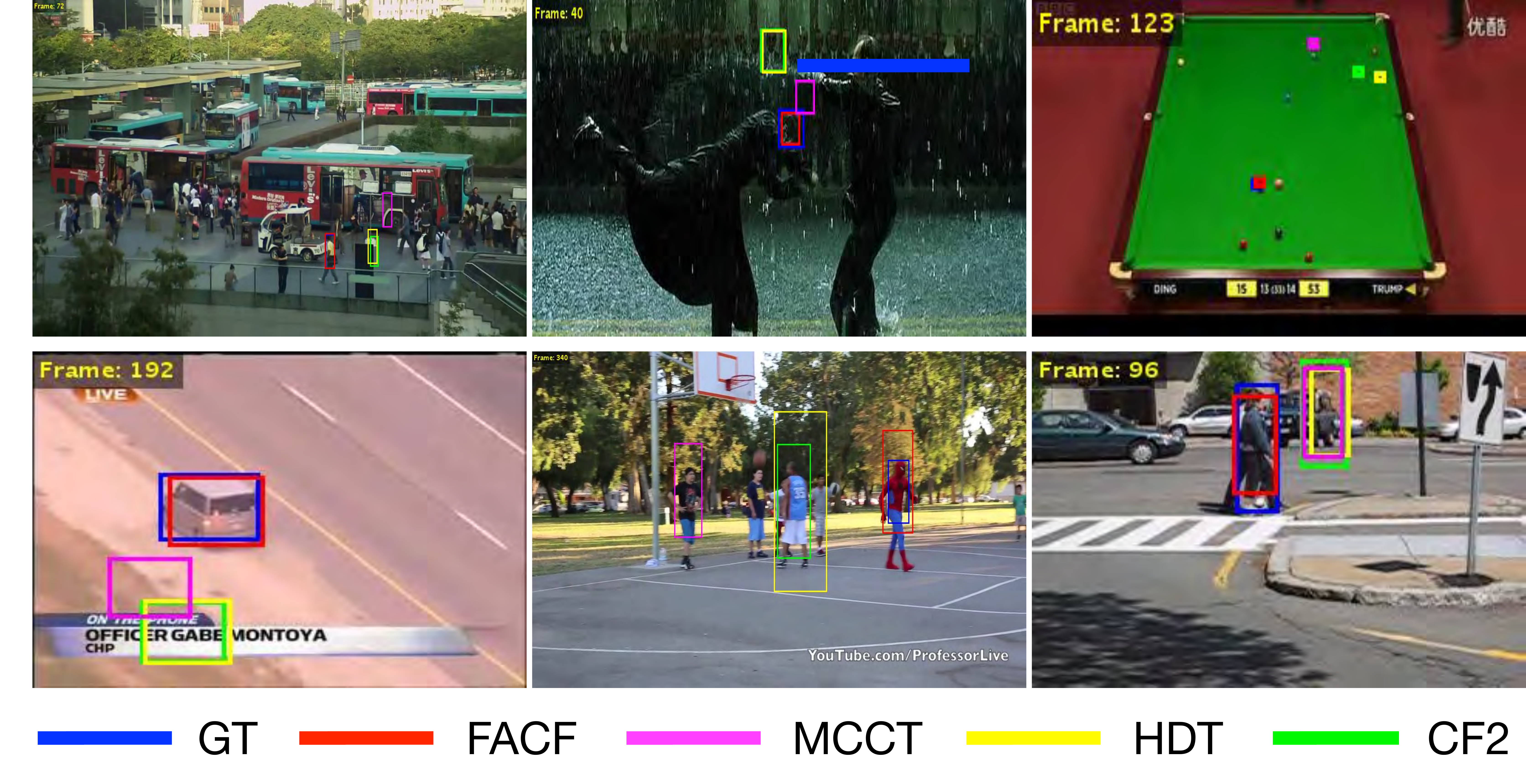}
\end{center}
   \caption{Comparison of the proposed tracker (FACF) with the ensembled trackers: MCCT \cite{CVPR2018MCCT}, HDT\cite{Hedged} and CF2 \cite{HCF} on sequences \emph{Busstation\_ce1}, \emph{Matrix}, \emph{Pool\_ce3}, \emph{Carchasing\_ce1}, \emph{Spiderman\_ce} and \emph{Couple} from TempleColor \cite{TC} data set. }
\label{fig:fig1}
\end{figure}

Lately, Discriminative Correlation Filter (DCF) trackers \cite{CVPR2017ECO, TPAMI2014KCF} have sparked a lot of interest thanks to their impressive performance and high frame-per-second (fps). In general, DCF trackers learn a correlation filter online to localize the target in consecutive frames by minimizing a least-squares loss for all circular shifts of a training sample. Since the correlation operation can be performed in the Fourier domain to simplify computational complexity, DCF trackers usually have an advantage in tracking speed. Based on the DCF framework, many improved trackers have been proposed. Improvement strategies include the use of adaptive region scale \cite{ECCV2014SAMF, 2014DSST}, multi-dimensional features \cite{ECCV2014SAMF, HCF}, long-term tracking \cite{Long1, MUSTer}, part-based tracking \cite{part1, part2}, end-to-end tracking \cite{CREST, end2}, and improved objective functions \cite{CVPR2017stapleca, filter1, filter2, filter3}. Moreover, the DCF framework incorporated with features extracted in deep convolutional neural networks (CNNs) has demonstrated state-of-the-art performance due to the powerful feature learning capabilities of CNNs. Recently, there are increasing interests in ensembled trackers. For example, Wang $et~al.$ proposed a Multi-Cue Correlation filter based Tracker (MCCT \cite{CVPR2018MCCT}) where multiple experts are constructed through DCF and each of them tracks the target independently by combining HOG \cite{HOG} features and VGG-19 (a type of CNN) \cite{VGG19} features ($L28$, $L37$). The suitable expert is selected for tracking in each frame through a just-the-right decision mechanism.
\begin{figure*}[t]
\begin{center}
\includegraphics[width=1.0\linewidth]{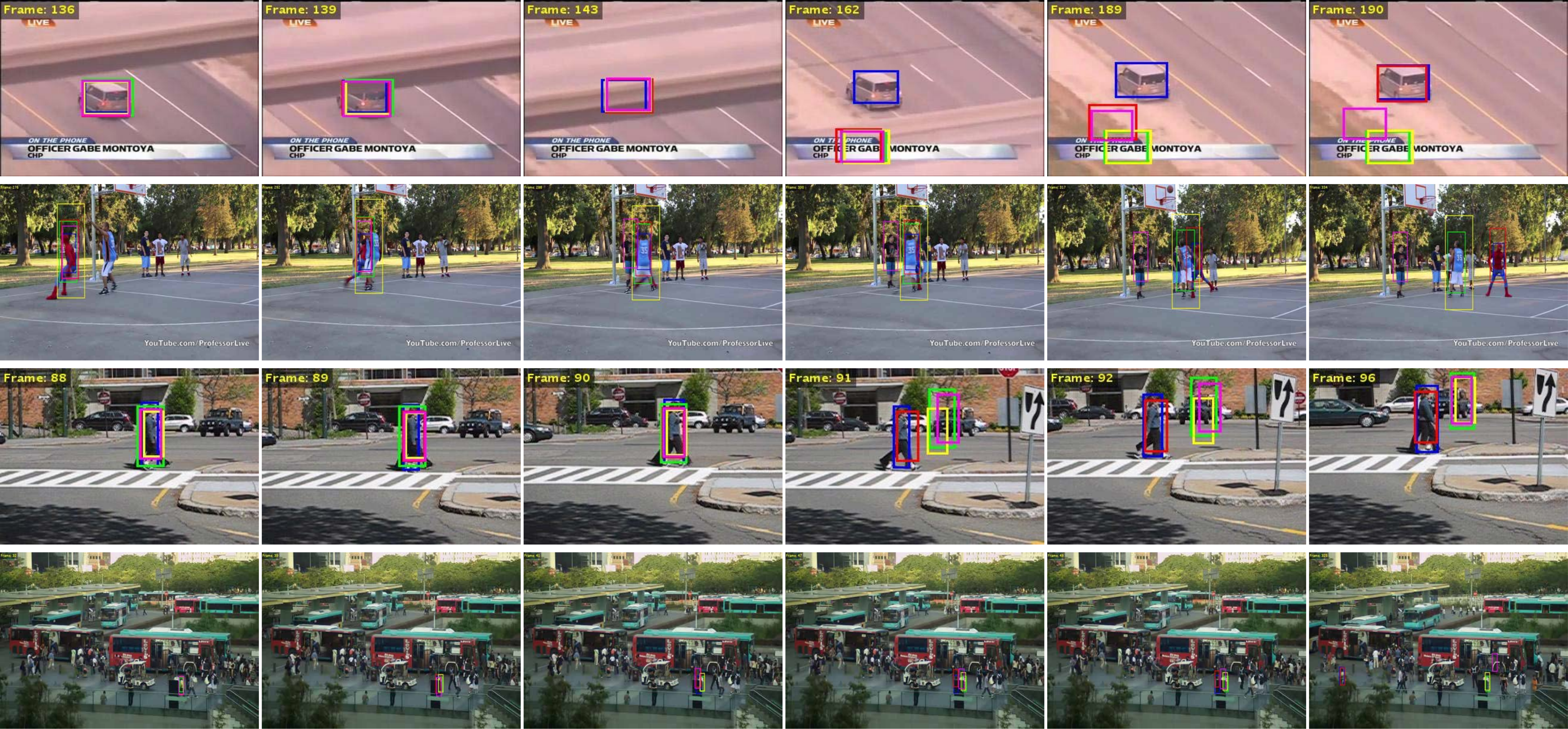}
\end{center}
   \caption{Tracking visualizations in successive frames among ensembled trackers on sequences \emph{Busstation\_ce1}, \emph{Carchasing\_ce1}, \emph{Spiderman\_ce} and \emph{Couple} from data set TempleColor \cite{TC}. The blue box indicates ground truth (GT), the red box indicates the proposed method (FACF), the magenta box indicates MCCT \cite{CVPR2018MCCT}, the yellow box indicates HDT \cite{Hedged} and the green box indicates CF2 \cite{HCF}.}
\label{fig:fig2}
\end{figure*}
There are many excellent ensembled trackers based on feature-level fusion \cite{CVPR2018MCCT, HCF, Hedged, ML4}, however, it is still a challenge problem to assign different features. For example, the high initial weight of high-level features in fusion trackers in \cite{HCF, Hedged, ML4}, results in the dominant role of semantic features in general. However, when encountering occasional misguided semantic information, a transient drift or wrong prediction will be amplified by insufficient online updates. These methods sometimes fail to fully explore the relationship among multiple features. In MCCT \cite{CVPR2018MCCT}, fusion strategy is achieved by experts, each of them is a linear combination of different level feature responses. In consecutive frames, all experts have a prediction result, and the best one is selected by a robust evaluation system. Although MCCT has gained huge performance improvement compared with the aforementioned ensembled trackers, more detail/suitable features are necessary to fully represent the object when the scene is complex as shown in Fig. \ref{fig:fig1} where different objects are overlapped and the target is confused with the other objects. With simplex or deficient features, the trackers may fail to track the correct target. Moreover, since all experts in each frame participate in tracking, it is easy to cause over-fitting and the missed target will never be re-tracked again (Fig. \ref{fig:fig2}), especially when background becomes clutter or occlusion emerges.

To tackle the problems above, in this paper, a novel hierarchical feature-aware tracking framework based on correlation filter is proposed (FACF). On the one hand, the proposed method maintains more experts (additional layers $L5$, $L10$, $L19$ in VGG-19 are considered) to learn more appropriate appearance models from more diverse views compared with existed ensembled trackers. Here, an individual expert is composed of a certain combination of feature responses. On the other hand, we propose a pre-known expert selection strategy that allows the system pre-assign suitable experts (Executive Experts) according to their performance in previous frames to mitigate overfittness existed in after-descision ensemble trackers (MCCT) and meanwhile keep computational efficiency. Due to the use of more significative features and the features are adaptively assigned to appropriate scenes, FACF is able to outperform existing emsembled trackers. As the example shown in Fig. \ref{fig:fig2}, FACF not only accurately tracks the target in complex scenes (especially on cluttered background or when occlusion emerges) but also re-track the correct target after it missed the target (first scene). Experiments on OTB-2013 \cite{2013OTB50}, OTB-2015 \cite{2015OTB100}, TempleColor \cite{TC} and VOT2017 \cite{VOT2017} demonstrate the effectiveness of the proposed framework, the superiority of the proposed method over compared ensembled trackers and the state-of-the-art performance of the proposed method.

The rest of this paper is organized as follows. Related background and motivation of FACF are discussed in Sec. II. Sec. \ref{cueacf} describes how to construct the scheme of FACF tracker by introducing additional efficient CNN's layers and ensembling a novel pre-known expert-adaptive selection strategy based on DCF tracking framework in detail. Sec. \ref{experiment} reports comparative experimental results on several publicly data sets. Finally, conclusions are drawn in Sec. \ref{conclusion}.


\section{Related Work}
In this section, we first give a brief discussion on two categories of trackers closely related to the proposed tracker: Correlation Filter (CF) based trackers and ensembled trackers, and then summarize contributions of the proposed method.
\subsection{CF based Trackers}Since Bolme $et~al.$ proposed the MOSSE \cite{CVPR2010MOSSE} which is the tracker using minimum output sum of squared error filter, CF framework have been studied as a robust and efficient approach in visual tracking. Major improvements to MOSSE have been made either ranking at the top of the benchmarks or remaining computationally efficient. Heriques $et~al.$ proposed a tracker CSK \cite{ECCV2012CSK} to exploit the circulant structure of the training patches and then proposed a tracker KCF \cite{TPAMI2014KCF} to train correlation filter in a kernel space with HOG features. Qi $et~al.$ proposed the tracker HDT \cite{Hedged} that fuses several DCFs through adaptive hedged method \cite{Hedmethod}. Luca $et~al$. proposed the tracker Staple \cite{CVPR2016STAPLE} which combines DCF and color histogram based model. Lukezic $et~al$ proposed the tracker CSR-DCF \cite{CSR-DCF} which constructs DCF with channel and spatial reliability. Danelljan $et~al.$ proposed the tracker SRDCF \cite{SRDCF} to alleviate the boundary effects by penalizing correlation filter coefficients depending on spatial location, and then proposed the tracker SRDCFdecon \cite{SRDCFdecon} to reduce the influence of corrupted samples in SRDCF. Recently, Danelljan $et~al.$ proposed the tracker C-COT \cite{ECCV2016C-COT} which adopts a continuous-domain formulation of the DCF and latter proposed ECO \cite{CVPR2017ECO}, the enhanced version of C-COT, to improves both speed and performance in \cite{ECCV2016C-COT} by introducing several efficient strategies.

\begin{figure}[t]
\begin{center}
\includegraphics[width=1.0\linewidth]{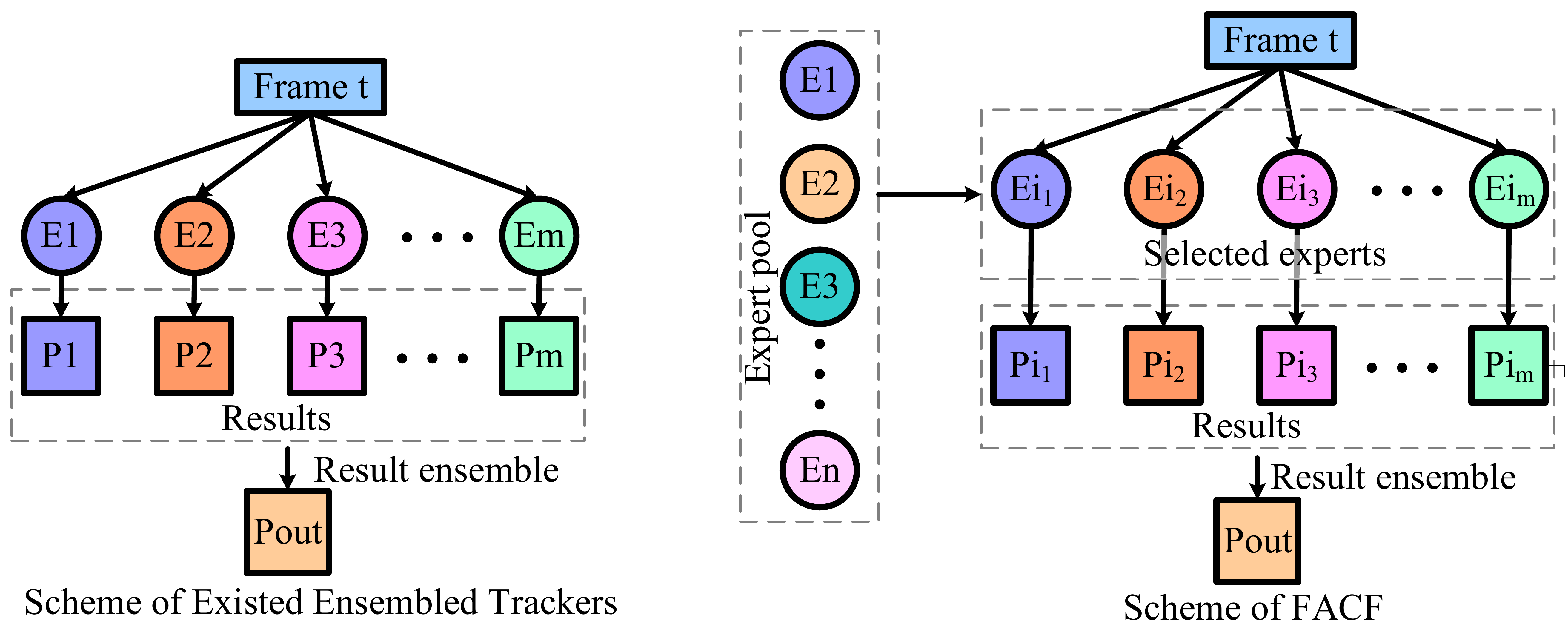}
\end{center}
   \caption{Difference between existing ensembled trackers and FACF. E denotes experts and P denotes the tracking result. Here n $\gg$ m.}
\label{fig:fig9}
\end{figure}
\subsection{Ensembled Trackers}
Recently, inspired from the mechanism of ensemble learning that combines multiple learners for decision, many ensemble trackers have been developed by combining multiple component trackers \cite{Ensemble1, Ensemble2, Ensemble3, Ensemble4, HCF, CVPR2018MCCT,Hedged}.
In some trackers, hand-crafted features are used \cite{Ensemble1, Ensemble2, Ensemble3, Ensemble4} to construct diverse component trackers. For example, ensemble approaches in \cite{Ensemble1, Ensemble3, Ensemble4} are based on the boosting framework \cite{framework1} and in them, each component weak tracker is incrementally trained to correctly classify the training samples which are missed by previous trackers. In \cite{Ensemble2}, the reliability of each component tracker as well as the final tracking result are inferred via a conditional particle filter. With the development of deep learning, the component trackers can be constructed by deep features \cite{Hedged, CVPR2018MCCT}. In \cite{Hedged}, tracking is taken as a decision-theoretic online learning task and the final tracking result is inferred by the expert decisions in multiple trackers.
In \cite{CVPR2018MCCT}, Wang $et~al$. considered not only feature-level fusion but also decision-level fusion to better explore the relationship of multiple features, and adaptively selects the expert that is suitable for a particular tracking task. Mechanisms of both trackers in \cite{CVPR2018MCCT} and \cite{Hedged} belong to afterthought tracking.

Although above methods have achieved state-of-the-art performance on several current public data sets, there is still room for improvement especially when there are cluttered background and occluded objects. Most of the CNN features used in these methods are concentrated on semantic-rich high-level features (extracted from top layers), and the details-rich layers (shallow and middle layers) are not well utilized. In theory, the objects will be more sufficiently represented with more types of features. But this will lead to exponential increase of computational complexity. In addition, the ensembled trackers are after-the-fact decision in the selection of the adaptive expert which results in misleading of inappropriate experts to some degree. Meanwhile, all the experts are trained by the same samples which may lead to overfitting. As a consequence, on the one hand, we consider not only high-level features but also shallow and middle level features to better explore the relationship of multiple features. On the other hand, we propose a novel pre-known expert selection strategy that adaptively selects the suitable experts in advance according to their performance in past frames to avoid misleading experts while keep computational efficiency.

\subsection{Contributions} The main contributions of our work can be summarized as follows. (1) We propose an novel Hierarchical Feature-Aware tracking framework (FACF) based on CF, which fuses features of multi-types (HOG and CNNs) and multi-layers (layers $L5$, $L10$, $L19$, $L28$ and $L37$ in VGG-19). The features we use are more sufficient than that in existing ensembled trackers in order to appropriately represent the target and avoid distraction from background. (2) We propose a novel pre-known expert-adaptive selection strategy that allows system to determine which experts are suitable for tracking in the next frame in advance. The performance of each expert is recorded during the consecutive frames. The experts with higher performance are more likely to be selected to execute in the following frames. The difference between FACF and existing ensembled trackers are illustrated in Fig. \ref{fig:fig9} where experts are first pre-selected from a large expert pool in FACF instead of fixed experts during tracking in other ensembled trackers. Extensive experiments demonstrate the effectiveness of feature-level-fusion method and pre-known expert-adaptive selection strategy used in FACF tracker. In addition, the state-of-the-art performance is also achieved on several challenging benchmarks. 

\section{The Proposed Method}
\label{cueacf}
\begin{figure*}[t]
\begin{center}
\includegraphics[width=1.0\linewidth]{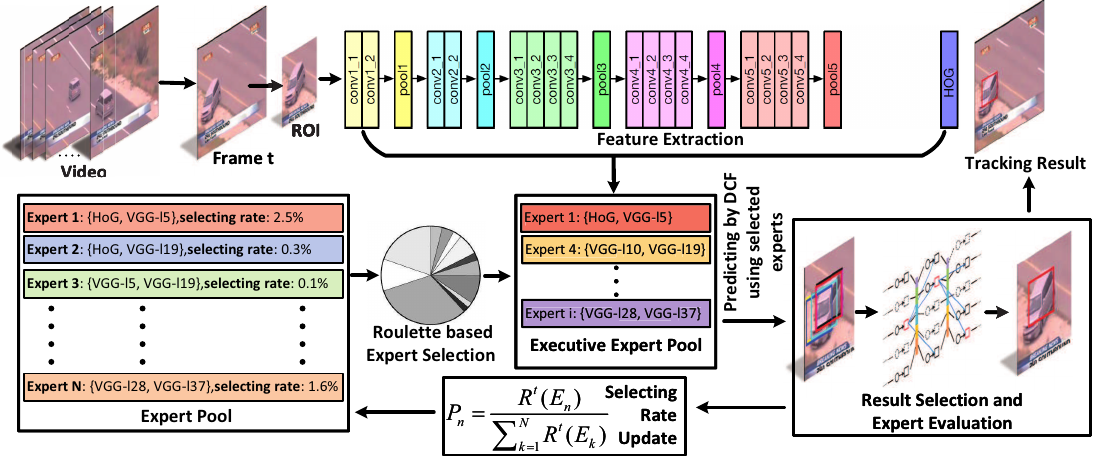}
\end{center}
   \caption{The framework of FACF. First, a set of executive experts is selected from the expert pool and the features are extracted and combined to train the executive experts. Then, each executive expert give its tracking result and they are evaluated by the fitness degree. Finally, the final result is selected and the selecting rate of each expert in the expert pool is updated according to their fitness. }
\label{fig:cueacf}
\end{figure*}
Fig. \ref{fig:cueacf} shows the scheme of the proposed tracker. Our approach uses the standard DCF \cite{CVPR2010MOSSE, TPAMI2014KCF} framework as the tracking framework and combines multiple features (in feature pool) to generate initial expert pool. However, if all the experts in the pool are executed in each frame, some inappropriate experts may mislead the result to degrade. Therefore, an executive expert pool is established which is composed of selected experts which is suitable in the current scene for each frame. The executive experts are selected according to a novel pre-known expert-adaptive selection strategy in order to avoid overfitting to some degree. Therefore, in this section, before the detailed discussion of our proposed method and for completeness, we first revisit the details of the DCF tracking framework \cite{CVPR2010MOSSE, TPAMI2014KCF}, which is the tracking framework of the proposed tracker (FACF), then introduce the composition of executive expert pool, the process of the best tracking result selection strategy, and the implementation details in turn.
\subsection{DCF Tracking Framework}
\label{tracking framework}
The goal of the standard DCF based tracker \cite{TPAMI2014KCF} is to learn a discriminative correlation filter $\bm{\omega}$ that can be applied to the region of interest (ROI) to infer the location of the target (i.e. location of maximum filter response) in consecutive frames. Given sample in the current frame, $\bm{\omega}$ is updated by optimizing the following regression problem:
\begin{equation}
\label{dcf}
\min_{\bm{\omega}}\|\bm{X\omega}-\bm{y}\|_2^2+\lambda\|\bm{\omega}\|_2^2
\end{equation}
where the training sample $\bm{X}$ is generated by all circular shifts of the patch $\bm{x}(m, n) \in \{0, 1, ..., M-1\} \times \{0, 1, ...N-1\}$, centered around the target, with Gaussian function label $\bm{y}(m,n)$ in terms of the shifted distance and $\lambda$ is a regularization parameter. The solution to filter $\bm{\omega}$ in the Fourier domain is given by \cite{TPAMI2014KCF}
\begin{equation}
\label{w}
\hat{\bm{\omega}}_d^* = \frac{\hat{\bm{x}}_d^*\odot \hat{\bm{y}}} {\sum_{i=1}^D\hat{{\bm{x}}}_i^*\odot \hat{\bm{x}}_i +\lambda}
\end{equation}
where $\hat{}$ denotes the Discrete Fourier Transform (DFT) (e.g., $\hat{\bm{x}} = F(\bm{x})$), $\odot{}$ is the element-wise product, $^*$ represents the complex-conjugate transformation and $d$ is the $d$-th $(d \in \{1, ..., D\})$ feature channel. In the next frame, the response $\bm{R}$ of a candidate patch $\bm{z}$ is computed following Eq. (\ref{R}). The target is identified as the candidate patch with the maximum value of $\bm{R}$.
\begin{equation}
\label{R}
\bm{R}= F^{-1}(\sum_{d=1}^D\hat{\bm{\omega}}_d \odot \hat{\bm{z}}_d^*)
\end{equation}
where $F^{-1}$ donates the Inverse DFT.
\subsection{Executive Expert Pool}
For completeness, before introducing the Executive Experts Pool, we first introduce the Feature Pool and the Initial Expert Pool adopted in our work.
\subsubsection{Feature Pool and Initial Expert Pool}
\label{feaselexpcom}
DCF is a basic framework where the target is represented by features and various types of feature extractors can be adopted. The effect of different features is also distinct. Hand-crafted features are efficient to capture low-level details but not adaptive. Deep learned features are captured by hierarchical architectures which can be learned from large scale datasets. The deep architecture can capture both detail-aware (from low-level layers) and semantic-aware (from high-level layers) features which is more adaptive. As discussed in HCF \cite{HCF}, features extracted from a single layer of a hierarchical architecture cannot result in accurate tracking. Therefore, HCF generates several DCF response maps via features from different layers simultaneously and performs a coarse-to-fine search from these maps.

In the proposed method, HOG \cite{HOG} is used to extract hand-crafted low-level features. As for CNN features, we remove the fully-connected layers and extract the outputs of the $L5$, $L10$, $L19$, $L28$ and $L37$ layers of VGG-19 \cite{VGG19} (Table \ref{IEP} lists the corresponding convolutional layers) as additional low-level, middle-level and high-level features, respectively. Due to the change of the target state during the tracking process, we propose an adaptive method to select the appropriate feature constructed experts (Executive Expert Pool) in order to represent a specific target. Note that since the features are adaptively selected, the feature pool we construct is richer and the number of features in it has almost no influence on computational efficiency. Meanwhile the more abundant feature pool can construct experts who are more suitable to the specific situation. In this paper, the feature pool consists of six types of features (HOG, $L5$, $L10$, $L19$, $L28$ and $L37$) and they are optionally combined into $C_6^1 + C_6^2 + C_6^3 + C_6^4 + C_6^5 + C_6^6 = 63$ experts (Initial Expert Pool). It deserves to be noted that some experts only consist of features from single layer. Although these experts may be less robust compared to other experts which are composed of features of multiple layers, diversity of tracking results provided by them is crucial in ensembled trackers \cite{43}.
\begin{table}[t]
\caption{Features and corresponding convolutional layers in VGG-19.}
\renewcommand\tabcolsep{7.0pt}
\renewcommand\arraystretch{1.2}
\begin{center}
\begin{tabular}{|c|c|c|c|c|c|}
\hline
~               &$L5$   &$L10$   &$L19$  &$L28$  &$L37$       \\
\hline
\hline
VGG-19 &conv1\_2 &conv2\_2 &conv3\_4 &conv4\_4 &conv5\_4\\
\hline
\end{tabular}
\end{center}
\label{IEP}
\end{table}
\subsubsection{Executive Expert Pool}
Usually, the sequences vary little between continuous frames, whether it is a target portion or a background portion. Therefore, we can infer the features that are suitable for representing the target in the next frame based on the performance of the expert in the previous frames.

\begin{figure}[t]
\begin{center}
\includegraphics[width=1.0\linewidth]{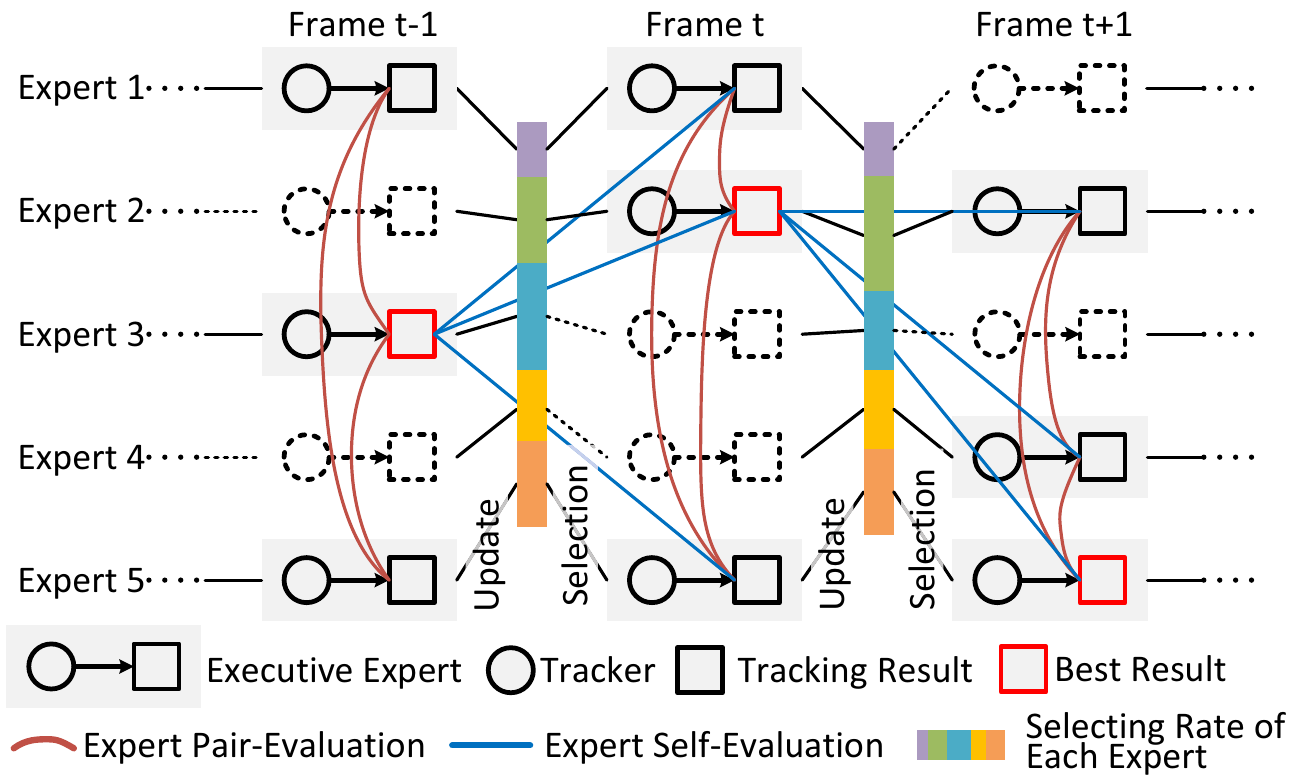}
\end{center}
   \caption{Graph illustration of FACF tracking in each frame. For clarity, only five experts are displayed and three experts are selected in each frame. }
\label{fig:expselcueacf}
\end{figure}

As shown in Fig. \ref{fig:expselcueacf}, before tracking, the executive experts are first selected according to their performance in previous frames evaluated by the fitness degree $R^t$. The greater the fitness degree $R^{t}(E_n)$ of the $E_n$ which is the $n$-th expert in the Expert Pool, the better the performance of $E_n$ in the past frames, and the sequence scene tends to vary less between adjacent frames. So, this expert should have a greater probability of being selected to perform the tracking tasks in the next frame. Note that if the expert does not perform tracking at the $t$-frame, the fitness degree maintains the same as in the $f-1$-th frame. Here, we use roulette-based selection strategy to select executive experts in the expert pool. This selection strategy has two benefits: on the one hand, each expert has the opportunity to be selected to perform the tracking in the next frame, which enriches the diversity of trackers; on the other hand, adaptively selecting suitable experts to perform the tracking can avoid misleading of inappropriate experts. The probability $P_n$ that the $E_n$ is selected into Executive Expert Pool in the next frame is computed as follows:
\begin{equation}
\label{F}
P_n = \frac{R^{t}(E_n)}{\sum_{n = 1}^N R^{t}(E_n)}
\end{equation}
\subsection{Best Result Selection and Fitness Update}
\label{eva}
Fig. \ref{fig:expselcueacf} shows the tracking flowchart of FACF in the $t$-th frame (5 experts for simplify). Before each frame is processed, executive experts are first selected according to the selecting rate in Eq. \eqref{F}. Then they track the target simultaneously and generate their tracking results respectively. Finally those results are evaluated via two criteria and the best one is selected as the final tracking result. Meanwhile, the selecting rate is updated according to the performance of each expert evaluated via the two criteria. After that, executive experts are trained and the process goes to the next frame.
Inspired by evaluation strategy used in \cite{CVPR2018MCCT}, in this paper, we also use the consistency with experts in both current and last frames to evaluate the performance of each executive expert which are: the consistency with best expert in the last frame represented by the blue lines in Fig. \ref{fig:expselcueacf}, and the consistency with other executive experts in the current frame represented by the brown lines in Fig. \ref{fig:expselcueacf}. Then, the two evaluation criteria are combined linearly to generate the fitness degree $R^t(E_n)$ of each expert $E_n$ in frame $t$.

The following mainly describes the computation of the fitness degree $R^t(E_n)$.
Firstly, since the executive experts are selected according to their performance, most of them is capable to track the target with high accuracy. Therefore, a good expert should be consistent with most other executive experts.
Let $E_1, E_2, ..., E_n, ..., E_{K}$ denote Expert $1$, Expert $2$, .., Expert $n$, ... , Expert $K$ in executive expert pool, respectively. Let $B_{E_n}^t$ donate the bounding box of $E_n$ in the $t$-th frame. The bounding box $B_{E_n}^t$ only contains the target state (e.g., location and scale) without any context information.
Then, we compute overlap ratios of the bounding boxes from different executive experts. The normalized overlap ratio $O_{(E_n, E_k )}^t$ between $E_n$ and $E_k$ at frame $t$ is calculated and represented by a nonlinear Gaussian function in order to reduce the gap between low and high ratios:
\begin{equation}
\label{O}
O_{(E_n, E_k)}^t = \exp\left[ - \left(1 - \frac{Area(B_{E_n}^t \cap B_{E_k}^t)}{Area(B_{E_n}^t \cup B_{E_k}^t)}\right)^2\right]
\end{equation}

The mean overlap ratio between $E_n$ and all the other executive experts is then obtained: $M_{E_n}^t = \frac{1}{K} \sum_{k = 1}^K O_{(E_n, E_k)}^{t}$.

In addition to the mean overlap ratio in this frame, it should also be stable in a short period.
Therefore, the fluctuation extent of overlap ratios in a $\Delta t$ (e.g., $5$) frames is given as follows:
\begin{equation}
\label{V}
V_{E_n}^t = \sqrt{\frac{1}{K} \sum_{k = 1}^K \left(O_{(E_n, E_k)}^{t} - \frac{1}{\Delta t} \sum_{\tau=1}^{\Delta t} O_{(E_n, E_k)}^{\tau+t-\Delta t}\right)^2}
\end{equation}
Then, to further avoid performance fluctuation of the experts, the weighted mean values of $M_{E_n}^t$ and $V_{E_n}^t$ are computed over $\Delta t$ frames with an increasing weight sequence $W = {\rho^0, \rho^1, ..., \rho^{\Delta t - 1}}, (\rho > 1)$: $\bar{M}_{E_n}^{t} = \sum_{\tau=1}^{\Delta t} W_{\tau} M_{E_n}^{\tau+t-\Delta t}/\sum_{\tau=1}^{\Delta t} W_{\tau}$ and $\bar{V}_{E_n}^{t} = \sum_{\tau=1}^{\Delta t} W_{\tau}V_{E_n}^{\tau+t-\Delta t}/\sum_{\tau=1}^{\Delta t} W_{\tau}$.

Finally, the pair-wise expert robustness score of $E_n$ at the $t$-th frame is defined as follows:
\begin{equation}
\label{Ri}
R_{pair}^t(E_n) = \frac{\bar{M}_{E_n}^{t}}{\bar{V}_{E_n}^{t} + \varepsilon}
\end{equation}
where $\varepsilon$ is a small constant that avoids the infinite pair-evaluation score for a zero denominator. A larger $R_{pair}^t(E_n)$ means better consistency with other executive experts and higher stability of the target state prediction.

Secondly, a good executive expert should also show reliability of its tracking results to those in previous frames as shown in Fig. \ref{fig:expselcueacf}. It is evaluated by the trajectory smoothness degree. The Euclidean distance measuring the shift between the previous best bounding box $B_{E_{best}}^{t - 1}$ and the current bounding box $B_{E_n}^{t}$ is first computed and represented by the nonlinear Gaussian function:
\begin{equation}
\label{S}
S_{E_n}^t = \exp \left[ -\frac{1}{2\sigma_{E_n}^2}\parallel c({B_{E_{best}}^{t - 1}}) - c({B_{E_n}^{t}}) \parallel^2\right]
\end{equation}
where $c({B_{E_n}^{t}})$ is the center of the bounding box $B_{E_n}^{t}$ and $\sigma_{E_n}$ is the average length of the width $W(B_{E_n}^t)$ and height $H(B_{E_n}^t)$ of the bounding box provided by $E_n$, i.e., $\sigma = \frac{1}{2}[W(B_{E_n}^t) + H(B_{E_n}^t)]$.
Similar to the pair-evaluation, we collect the previous movement information to consider the temporal stability. Finally, the self-evaluation score is defined by $R_{self}^t(E_t) = \sum_{\tau=1}^{\Delta t} W_{\tau}S_{E_i}^{\tau+t-\Delta t}/\sum_{\tau=1}^{\Delta t} W_{\tau}$. The higher $R_{self}^t(E_i)$ means the better reliability of the tracking trajectory.

The fitness degree $R^{t}(E_n)$ of the $E_n$ in $t$-th frame is a linear combination of its pair-evaluation score $R_{pair}^{t}(E_n)$ and self-evaluation score $R_{self}^{t}(E_n)$:
\begin{equation}
\label{r}
R^{t}(E_n) = \mu R_{pair}^{t}(E_n) + (1 - \mu) R_{piar}^{t}(E_n)
\end{equation}
where $\mu$ is the parameter to trade off the pair-evaluation and self-evaluation weights.
\subsection{Working Mechanism}
In the tracking process, we selected 6 feature layers (HOG, $L5$, $L10$, $L19$, $L28$ and $L37$) to form 63 experts ($C_6^1 + C_6^2 + C_6^3 + C_6^4 + C_6^5 + C_6^6 = 63$) which is much more than existing emsembled trackers (e.g., 7 experts in MCCT). Before each frame is processed, we adaptively select $K$ executive experts to perform tracking in this frame based on the robustness score of each expert in the expert pool. This robustness score comes from the robustness of its own trajectory on the one hand and the robustness of other experts trajectories on the other hand. Excellent experts with high scores are highly likely to be selected, and experts with low scores are less likely to be selected in this scheme. This adaptive learning scheme is robustness to deal with various scenarios.

In a simplex scene (e.g., on a simple background, stable motion, and slight deformation), experts who are good at dealing with corresponding scenes are with high fitness degree. Then they are more likely to select as the executive experts. As a consequence, when some experts are always selected, FACF degrades into MCCT where fixed experts are used. But the executive experts in FACF are adaptively selected instead of manually assigned. When the scene becomes complex (e.g., the background is blurred, occlusion occurs, and even the target transforms into different appearance), the fitness degree of the experts in the pool is approximately balanced due to the large divergence between them. Therefore, all experts are likely to be selected and the diversity of experts will lead to more robustness. In existing popular ensembled trackers, the experts are fixed and each expert are trained by the last frame. When the target changes a lot, those experts may be flustered to recognize the correct target. In FACF, since not all experts are selected in the last frame, executive experts in this frame are trained by different frames which can avoid overfitting. Moreover, when a target is missed by the tracker, other trackers may keep tracking the wrong target due to they are trained by the wrong samples. FACF is more likely to successfully re-track the correct target with multiple diverse experts.


\subsection{Implementation Details}
\label{modupd}

As in \cite{TPAMI2014KCF}, in order to avoid the boundary effects, we also apply the Hann window to the signals. On scale-adaptive estimation, we follow the DSST tracker \cite{DSST}. Besides, inspired by \cite{DCF}, in order to enhance the spatial reliability, the training sample is first masked by color information in a simple way: $\bm{X}^{'} = \bm{X} \odot \bm{C}$. Here $\bm{X}$ denotes the data matrix and $\bm{C}$ denotes the color mask which is generated via the histogram-based per-pixel score map \cite{CVPR2016STAPLE} of ROI. The online update of the numerator $\hat{\bm{A}}_{d}$ and the denominator $\hat{\bm{B}}_{d}$ in Eq. \eqref{w} with the $t$-th frame is listed as follows:
\begin{equation}
\begin{split}
\label{U}
\bm{A}_d^t &= (1-\eta)\hat{\bm{A}}_d^{t-1} + \eta~ \hat{\bm{y}}\odot \hat{\bm{x}}_d^{*t} \\
\bm{B}_d^t &= (1-\eta)\hat{\bm{B}}_d^{t-1} + \eta~\sum_{i=1}^D \hat{\bm{x}}_i^{*t} \odot \hat{\bm{x}}_i^t \\
\hat{\bm{\omega}}_d^{*t} &= \frac{\bm{A}_d^t}{\bm{B}_d^t + \lambda}
\end{split}
\end{equation}
where $\eta$ denotes the learning rate.
We follow the parameters in standard DCF framework \cite{TPAMI2014KCF} to construct experts.
Besides, all trackers are run on the same workstation (Intel Xeon CPU 643 E5-2630 v3 2.4GHz $\times$ 32, 64GB RAM) using Ubuntu 16.04, MATLAB R2017a.

\section{Experimental Verification}
\label{experiment}
\subsection{Data Sets and Evaluation Metrics}
\begin{figure}[t]
\begin{center}
\includegraphics[width=1.0\linewidth]{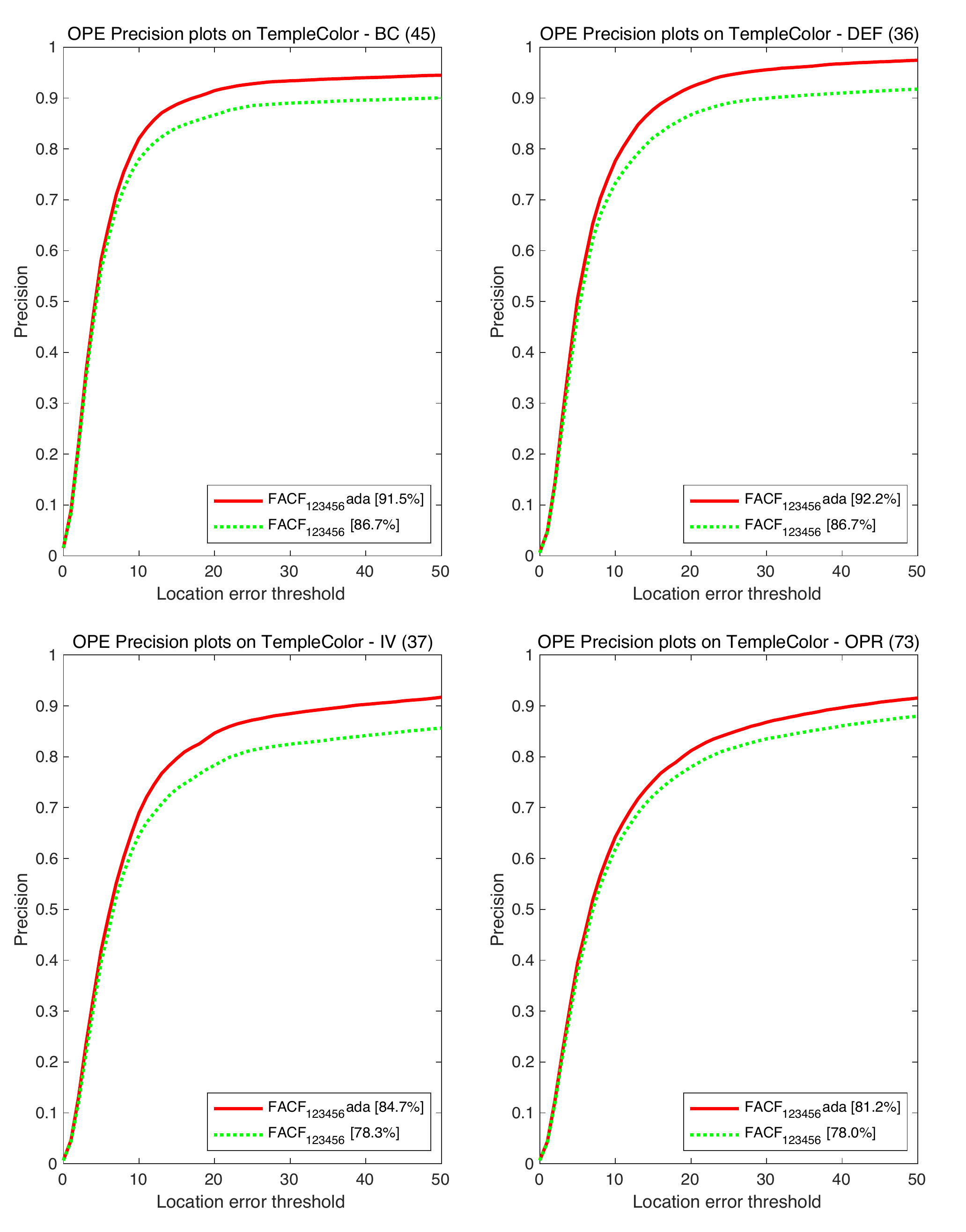}
\end{center}
   \caption{Precision plots on TempleColor for FACF\_{ada} and FACF\_{all}.  }
\label{fig:ada1}
\end{figure}

\begin{figure}[h]
\begin{center}
\includegraphics[width=1.0\linewidth]{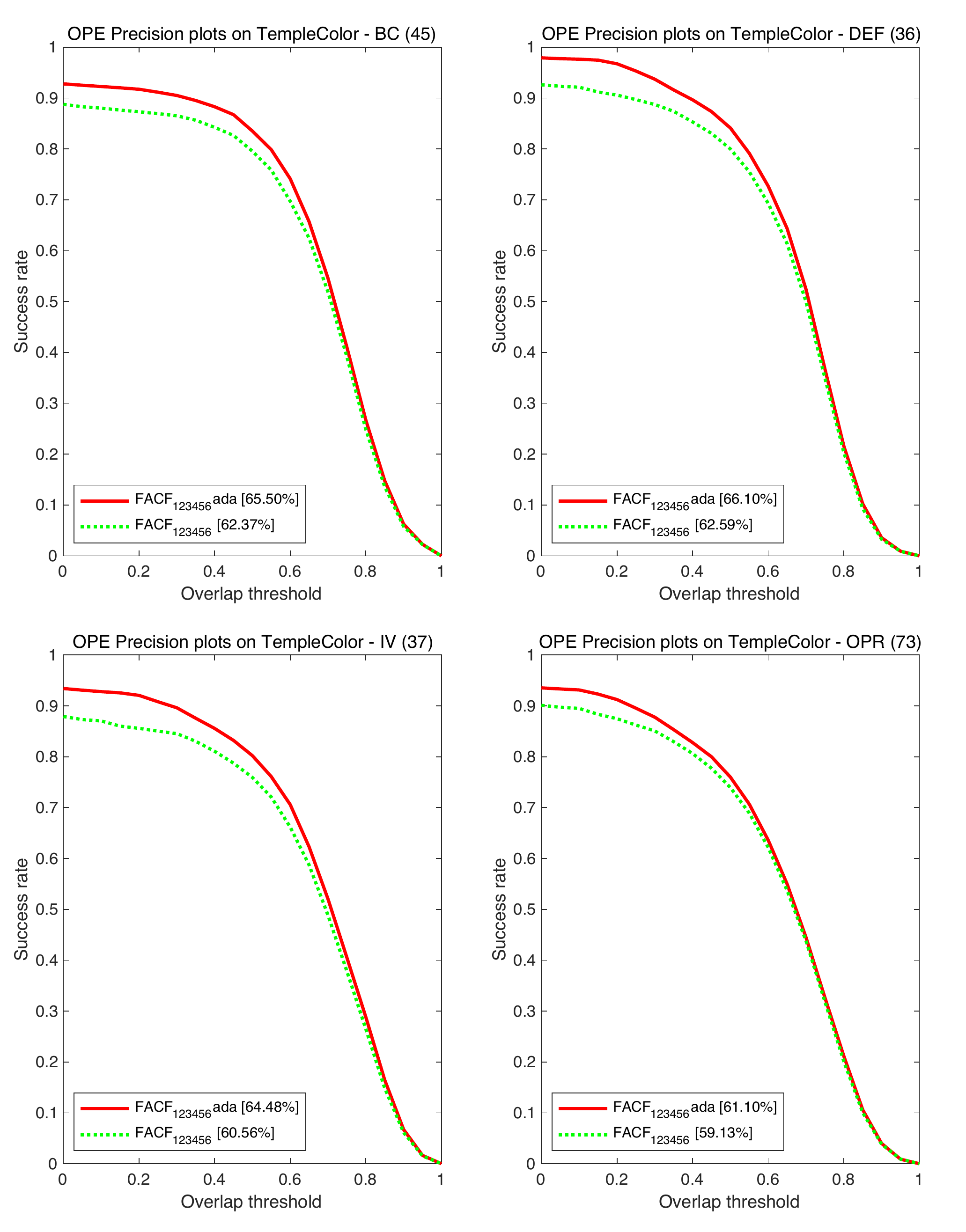}
\end{center}
   \caption{Success plots on TempleColor for FACF\_{ada} and FACF\_{all}.  }
\label{fig:ada2}
\end{figure}
Our method is evaluated on four representative benchmark data sets, i.e., OTB-2013 \cite{2013OTB50}, OTB-2015 \cite{2015OTB100}, TempleColor (color), and VOT2017 \cite{VOT2017}. OTB-2013 contains 50 sequences, OTB-2015 contains 100 sequences, TempleColor contains 129 sequences and VOT2017 contains 60 sequences. Both OTB-2015 and TempleColor have 11 attributes, namely Illumination Variation (IV), Scale Variation (SV), Occlusion (OCC), Deformation (DEF), Motion Blur (MB), Fast Motion (FM), In-Plane Rotation (IPR), Out-of-Plane Rotation (OPR), Out-of-View (OV), Background Blur (BC) and Low Resolution (LR). For OTB-2015 and TempleColor, a no-rest evaluation method: precision and success using one-pass evaluation (OPE) is used to evaluate the methods which is also used in \cite{2013OTB50, 2015OTB100}. Precision measures the error between centers of the tracked and ground truth bounding boxes. 
For precision, the compared trackers are ranked by the common threshold of 20 pixels (P(20px)). Success measures the intersection over union (IoU) between the tracked and ground truth bounding boxes. The success plot is obtained by the percentage of correctly predicted bounding boxes (y-axis) versus the varied overlap (x-axis). The area under the curve (AUC) is used to rank the compared trackers.
\begin{table}[t]
\caption{Performance comparison of different features used in FACF on TempleColor \cite{TC}. 1, 2, \textbf{3}, \textbf{4}, \textbf{5} and 6 respectively represents feature of hand-craft HOG \cite{HOG} and hidden layers of deep VGG-19 \cite{VGG19} (\emph{L5}, \emph{L10}, \emph{L19}, \emph{L28}, \emph{L37}). \emph{All} represents all experts are used in the corresponding tracker and \emph{Ada.} represents the adaptive expert selection strategy was used (the executive expert number of $28$ is used in \textbf{FACF}(1, \textbf{2}, \textbf{3}, \textbf{4}, 5, 6)\_{\emph{Ada.}}). }
\renewcommand\tabcolsep{11.5pt}
\renewcommand\arraystretch{1.2}
\footnotesize
\begin{center}
\begin{tabular}{|l|c|c|c|}
\hline
~                &Features          & P(20px)    & AUC       \\
\hline
\hline
MCCT(1, 5, 6) \cite{CVPR2018MCCT}              &\emph{All}                         & 80.0\%   & 58.98\% \\
\textbf{FACF}(1, \textbf{2}, 5, 6)                      &\emph{All}                          & 80.1\% & 58.79\%\\
\textbf{FACF}(1, \textbf{3}, 5, 6)                        &\emph{All}                         & 80.3\% & 59.58\%\\
\textbf{FACF}(1, \textbf{4}, 5, 6)                        &\emph{All}                       & 80.3\% & 59.13\%\\
\textbf{FACF}(1, \textbf{2}, \textbf{3}, 5, 6)            &\emph{All}                &80.7\% & 59.83\%\\
\textbf{FACF}(1, \textbf{2}, \textbf{4}, 5, 6)              &\emph{All}                      & 80.8\% & 59.80\%\\
\textbf{FACF}(1, \textbf{3}, \textbf{4}, 5, 6)               &\emph{All}                 & 80.5\% & 59.58\%\\
\textbf{FACF}(1, \textbf{2}, \textbf{3}, \textbf{4}, 5, 6) &\emph{All}   & \textbf{81.1\%} & \textbf{60.08\%}\\
\textbf{FACF}(1, \textbf{2}, \textbf{3}, \textbf{4}, 5, 6)&\emph{Ada.}    & \textbf{82.2\%} & \textbf{60.97\%}\\
\hline
\end{tabular}
\end{center}
\label{fc}
\end{table}

\begin{figure*}[h]
\begin{center}
\includegraphics[width=1.0\linewidth]{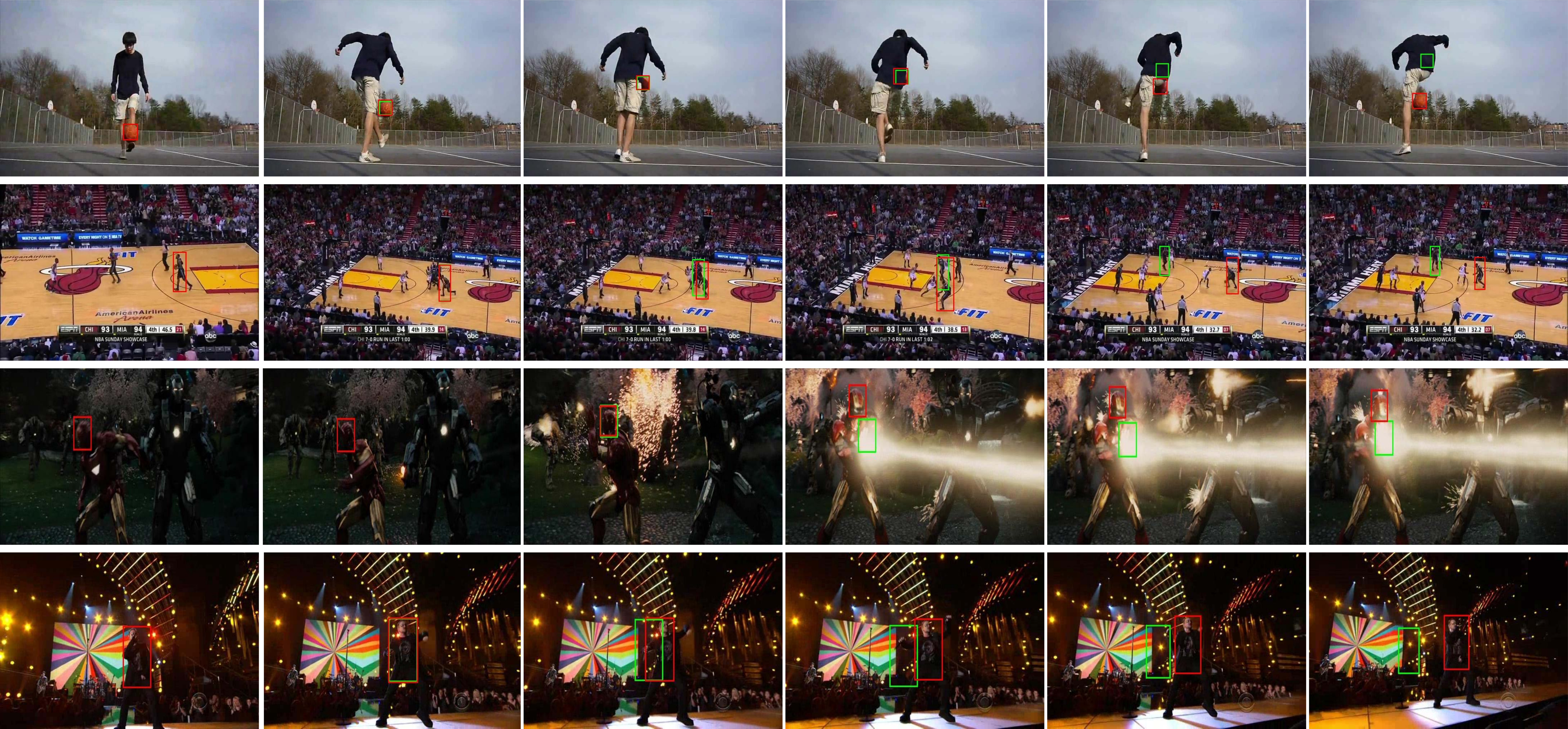}
\end{center}
   \caption{Visualization of tracking results in successive frames. The red box indicates FACF with adaptively expert selection strategy (FACF\_ada) and the green box indicates FACF without adaptively expert selection strategy (FACF\_{all}).  }
\label{fig:vis}
\end{figure*}

For VOT2017, the Expected Average Overlap (EAO),  Accuracy (Acc.) and Robustness (Rob.) are considered to evaluate the performance of trackers. EAO estimates the average overlap that the tracker is expected to achieve on short-term sequences which have the same visual properties as the given dataset.
The Acc. measures the average overlap between the predicted and ground truth bounding boxes over frames where the target is successfully tracked. The Rob. is the number of frames where the tracker misses the correct target. The values of these criteria in the experiments are the average values over 5 independent runs.


\subsection{Experiments on Framework Settings}

We select the data set TempleColor \cite{TC} with the most sequences (129) to verify the effectiveness of the proposed method with different settings. First, we test the effectiveness of different feature compositions. Second, we verify the impact of the number of execution experts on the tracking performance.
\subsubsection{Experiments on Feature Composition}
To demonstrate the effectiveness of feature composition in FACF, we compared the performance of FACF variants with different features. MCCT \cite{CVPR2018MCCT} uses the HOG and features extracted from $L28$ and $L37$ layers of VGG-19 as the baseline (\textbf{FACF}(1, \textbf{2}, 5, 6)). FACF incorporates more features and the performance of them are listed in Table \ref{fc}. With more features, the learner could represent the tracked object more sufficiently. Therefore, FACF achieves better performance than the baseline, especially when all the features are incorporated. Moreover, FACF uses feature adaptive strategy to further improve the performance and the performance of them is exhibited via precision and success plots in Fig. \ref{fig:ada1} and \ref{fig:ada2}, respectively. 

\begin{figure}[t]
\begin{center}
\includegraphics[width=1.0\linewidth]{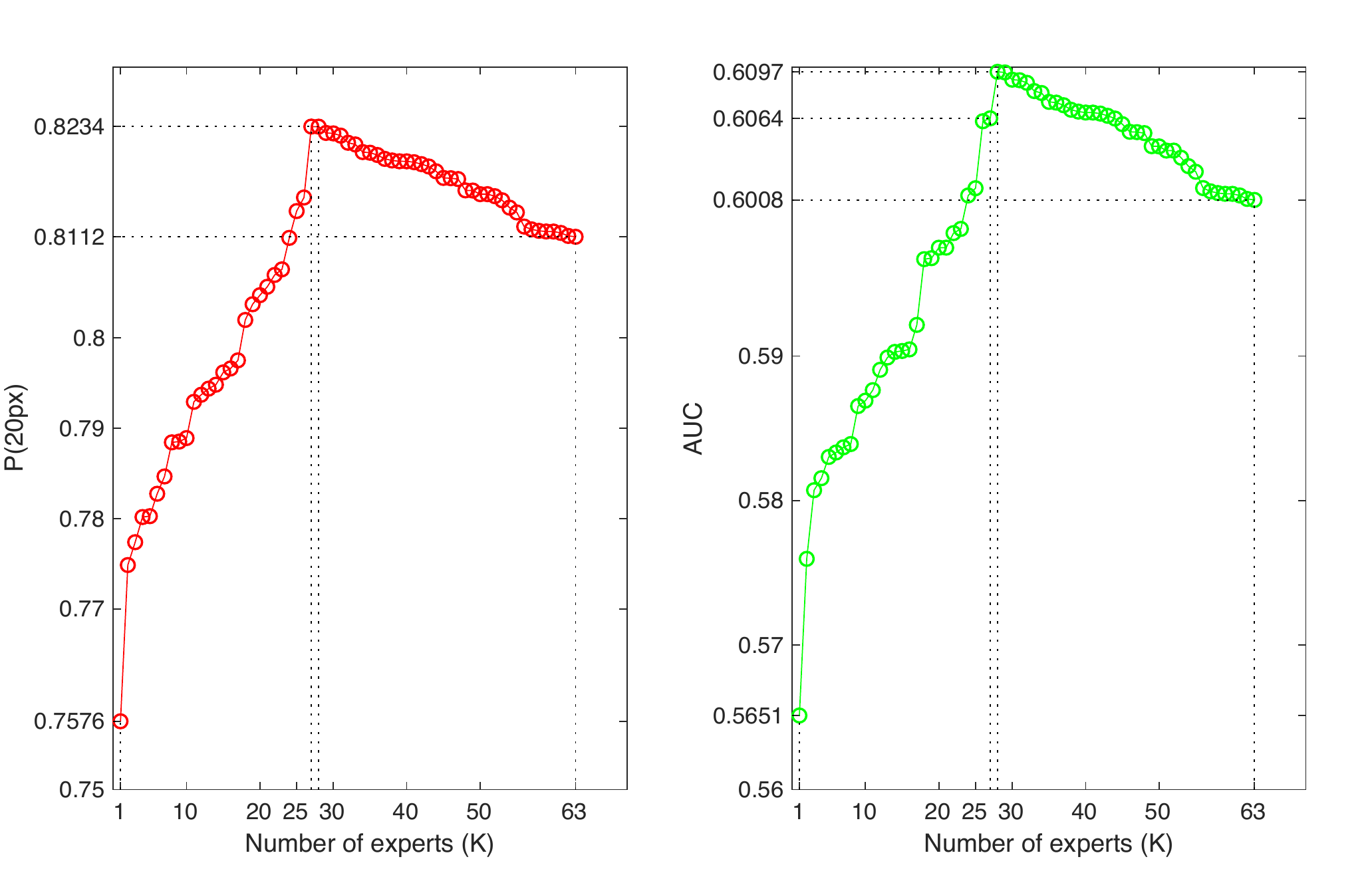}
\end{center}
   \caption{Performance of P(20px) and AUC with different number of executive experts on TempleColor \cite{TC}.  }
\label{fig:spp}
\end{figure}

In summary, our method with adaptive selection strategies (FACF\_ada) performs better on P(20px) with 1.1\% and on AUC with 0.89\% than the method without adaptive selection strategies (FACF\_all) (Fig. \ref{fig:ada1} and Fig. \ref{fig:ada2}). Moreover, the performance improvement of FACF\_ada is more obvious than FACF\_all on some specific attributes, for example, for the attribute of BC, the improvement is 4.8\% on P(20px) and 3.13\% on AUC, for DEF, it is 5.5\% on P(20px) and 3.51\% on AUC, for IV, it is 6.4\% on P(20px) and 3.92\% on AUC, and for OPR, it is 3.2\% on P(20px) and 1.97\% on AUC. This further demonstrates the effectiveness of the proposed framework, especially on attributes of BC, DEF, IV, and OPR. As Fig. \ref{fig:vis} shows, FACF\_ada is more robust than FACF\_all in some complex situations. This is mainly due to the fact that in some complicated situations, FACF\_ada can adaptively select the appropriate experts to perform tracking.

\begin{figure}[t]
\begin{center}
\includegraphics[width=1.0\linewidth]{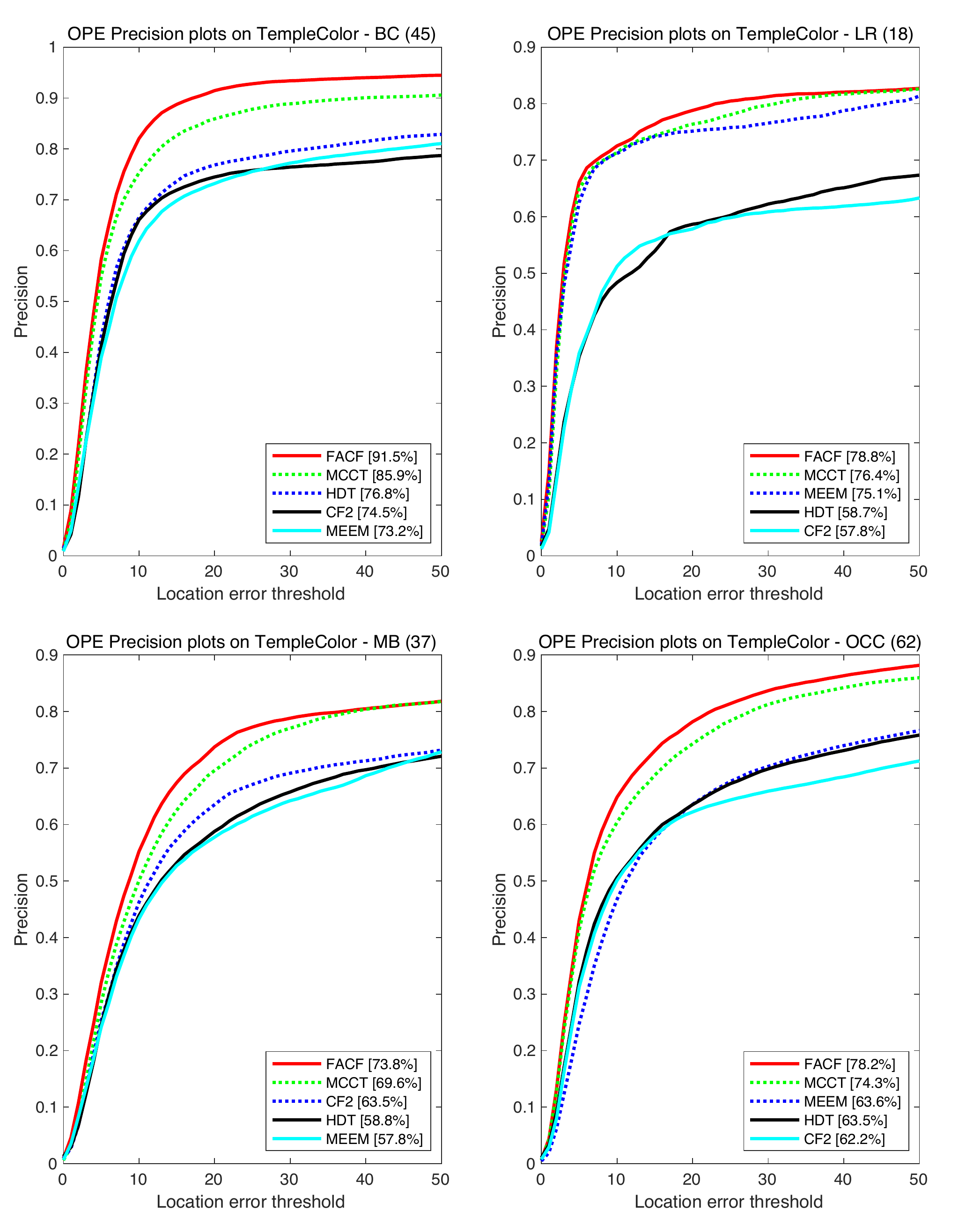}
\end{center}
   \caption{Precision plots on TempleColor for 4 attributes among ensembled trackers. }
\label{fig:3}
\end{figure}

\begin{figure}[t]
\begin{center}
\includegraphics[width=1.0\linewidth]{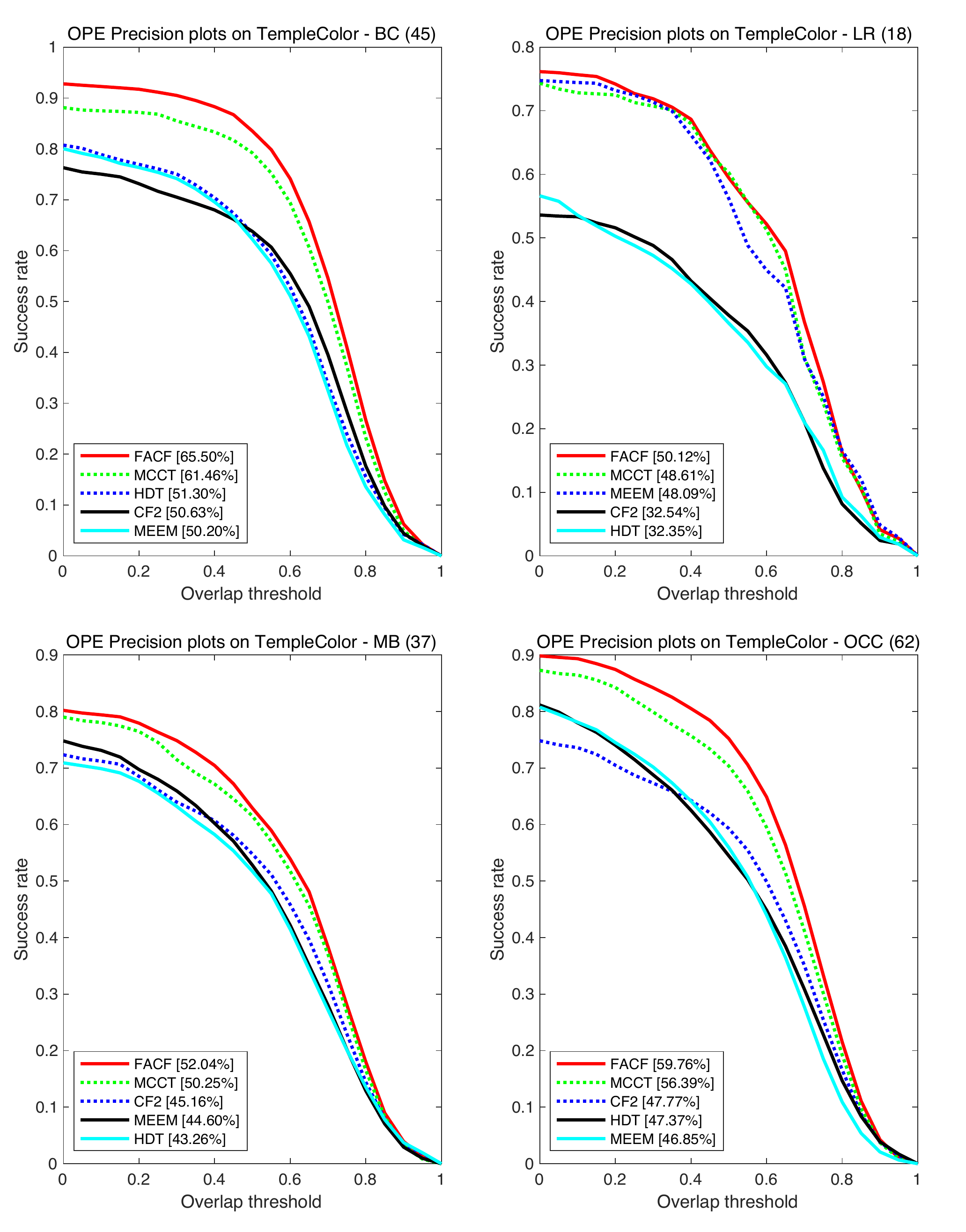}
\end{center}
   \caption{Success plots on TempleColor for 4 attributes among ensembled trackers. }
\label{fig:4}
\end{figure}

\newcommand{\tabincell}[2]{\begin{tabular}{@{}#1@{}}#2\end{tabular}}

\begin{table*}[h]
\caption{Performance comparison on P(20px)($\%$) and AUC($\%$) with several state-of-the-art ensembled trackers on benchmarks TempleColor \cite{TC}, OTB-2013 \cite{2013OTB50}, and OTB-2015 \cite{2015OTB100}. }
\renewcommand\tabcolsep{3.7pt}
\renewcommand\arraystretch{1.2}
\begin{center}
\begin{tabular}{|l|c|c|c|c|c|}
\hline
P(20px)/AUC&FACF(\textbf{Ours})&\tabincell{c}{MCCT~\cite{CVPR2018MCCT}\\2018~(CVPR)}&\tabincell{c}{CF2~\cite{HCF}\\2016~(ICCV)}&\tabincell{c}{HDT~\cite{Hedged}\\2016~(CVPR)}&\tabincell{c}{MEEM~\cite{ECCV2014MEEM}\\2014~(ECCV)}    \\
\hline
\hline
TempleColor \cite{TC}& \textbf{82.2\%}/\textbf{60.97\%}  &80.0\%/58.98\%   &69.0\%/50.01\%  &68.3\%/48.19\% & 70.7\%/50.43\%\\
OTB-2013 \cite{2013OTB50}& \textbf{93.2\%}/\textbf{71.04\%} &92.6\%/70.89\%   & 88.5\%/63.92\%  &86.3\%/58.81\% &81.2\%/56.74\% \\
OTB-2015 \cite{2015OTB100}& \textbf{91.9\%}/\textbf{69.44\%}   & 91.4\%69.08\%  & 84.6\%/61.14\%  &  84.4\%/56.74\% & 77.4\%/53.68\%\\

\hline
\end{tabular}
\end{center}
\label{fig:et}
\end{table*}

\subsubsection{Experiment on Number of Executive Experts}To analyze the impact of the number of executive experts on FACF, we compare the performance of FACF (1, \textbf{2}, \textbf{3}, \textbf{4}, 5, 6) under different numbers of selected experts. The results are exhibited with line graph as shown in Fig. \ref{fig:spp}. With increase of the number of experts, more experts are used to determine the position of tracked object. Therefore, the performance increases as well. However, when the experts are over 28, the performance begin to decrease. Because with more experts, the learner may overfit to the current frame. Therefore, when the tracked object is lost, it may not be recovered in the following frames. The performance is evaluated by P(20px) and AUC and they both can reflect the performance of FACF on some aspects. But with P(20px), the best performance is achieved when the number of experts is $27$ and that of AUC is $28$. In the following experiments, we select $28$ adaptive experts since AUC is more usually used to rank trackers.
\subsection{Comparison with Ensembled Trackers}To better demonstrate the effectiveness of the proposed method which is a type of ensembled tracker, we compare the best performed FACF tracker to several state-of-the-art integrated trackers including MCCT \cite{CVPR2018MCCT}, CF2 \cite{HCF}, HDT \cite{Hedged} and MEEM \cite{ECCV2014MEEM} on three public benchmarks TempleColor \cite{TC}, OTB-2013 \cite{2013OTB50}, and OTB-2015 \cite{2015OTB100}.

Table \ref{fig:et} lists the P(20px) and AUC values generated by the compared methods on the three benchmarks. The proposed tracker FACF performs best no matter on P(20px) or AUC. In particular, FACF increases by $\{2.2\%, 13.2\%, 13.7\%, 11.5\%\}$ on P(20px) and by $\{1.99\%, 10.96\%, 12.78\%, 10.54\%\}$ on AUC over trackers MCCT, CF2, HDT and MEEM on TempleColor \cite{TC}, respectively. It increases by $\{0.6\%, 4.7\%, 6.9\%, 12.0\%\}$ on P(20px) and by $\{0.15\%, 7.12\%, 12.23\%, 14.30\%\}$ on AUC over trackers MCCT, CF2, HDT and MEEM on OTB-2013 \cite{TC}, respectively. It increases by $\{0.5\%, 7.3\%, 7.5\%, 14.5\%\}$ on P(20px) and by $\{0.36\%, 7.94\%, 12.34\%, 15.40\%\}$ on AUC over trackers MCCT, CF2, HDT and MEEM on OTB-2015 \cite{TC}, respectively. This experiment demonstrates that the overall performance of FACF is significantly improved over existing ensembled trackers. Then we compare them under specific scenarios (different attributes) via precision and success plots as shown in Fig. \ref{fig:3} and \ref{fig:4}, respectively.


FACF achieves the best performance compared with other ensembled trackers on benchmarks containing various scenarios. While FACF achieves excellent performance on some specific attributes.
The most significant improvement is achieved in the cases of BC, LR, MB, and OCC (Fig. \ref{fig:3} and Fig. \ref{fig:4}). For BC, FACF exceeds on P(20px) with $\{5.6\%, 14.7\%, 17.0\%, 18.3\%\}$ and AUC with $\{4.04\%, 14.20\%, 14.87\%, 15.3\%\}$ compared with ensembled trackers MCCT \cite{CVPR2018MCCT}, CF2 \cite{HCF}, HDT \cite{Hedged} and MEEM \cite{ECCV2014MEEM}, respectively. For OCC, FACF exceeds on P(20px) with $\{3.9\%, 14.6\%, 14.7\%, 16.0\%\}$ and AUC with $\{3.37\%, 11.09\%, 11.39\%, 12.19\%\}$ compared with ensembled trackers MCCT \cite{CVPR2018MCCT}, CF2 \cite{HCF}, HDT \cite{Hedged} and MEEM \cite{ECCV2014MEEM}, respectively. Namely, if the object appearance changes drastically (OCC) or if the background looks similar to the target (BC), our framework is very beneficial no matter on P(20px) and AUC. This is largely due to the adaptively expert selection strategy used in FACF that allows tracker adaptively select the appropriate features to achieve tracking in this scenarios.

\subsection{Comparison with State-of-the-art Trackers}
\begin{figure}[t]
\begin{center}
\includegraphics[width=1.0\linewidth]{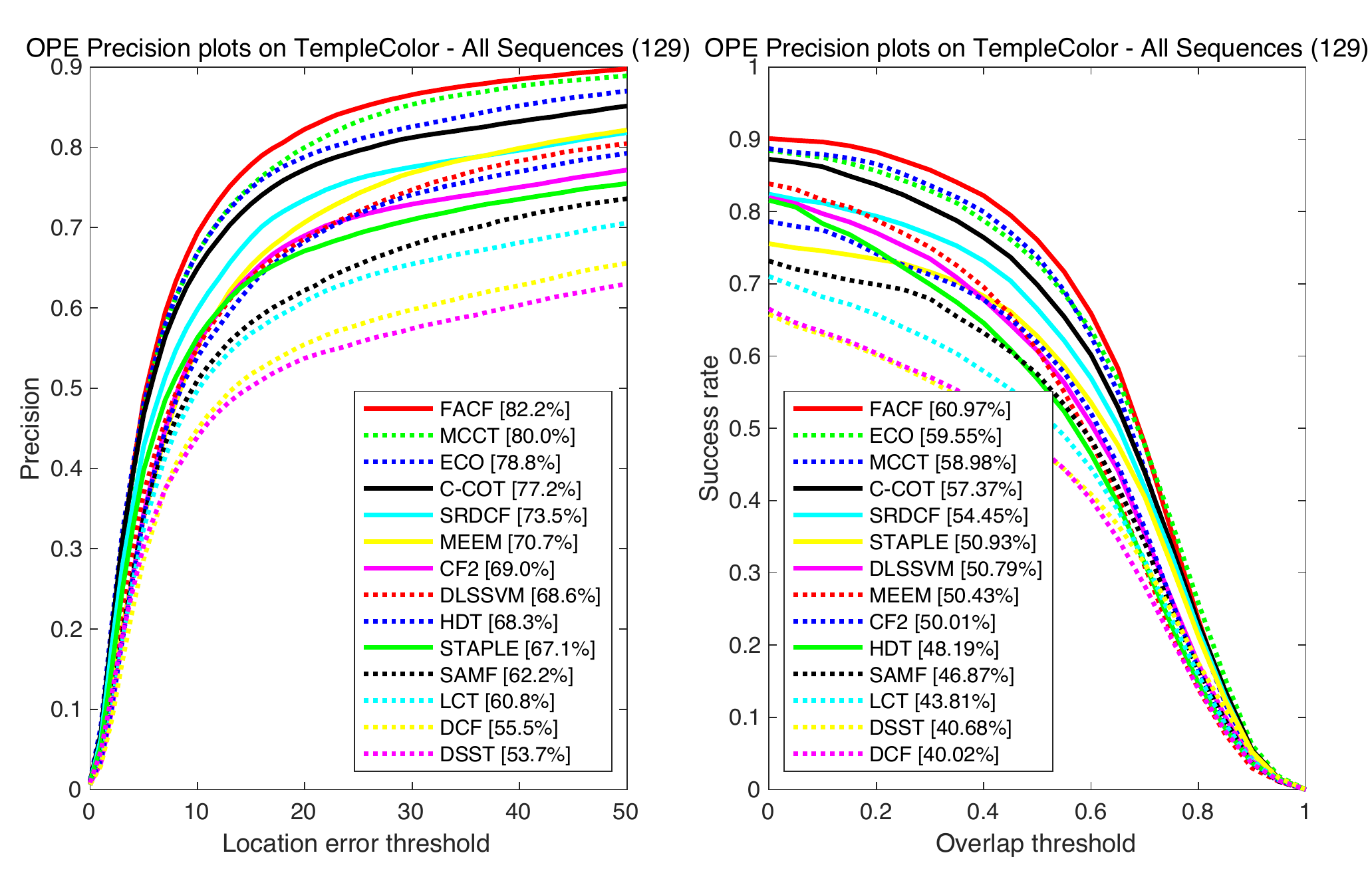}
\end{center}
   \caption{Precision and success plots on TempleColor \cite{TC}. }
\label{fig:tc}
\end{figure}
\begin{figure}[t]
\begin{center}
\includegraphics[width=1.0\linewidth]{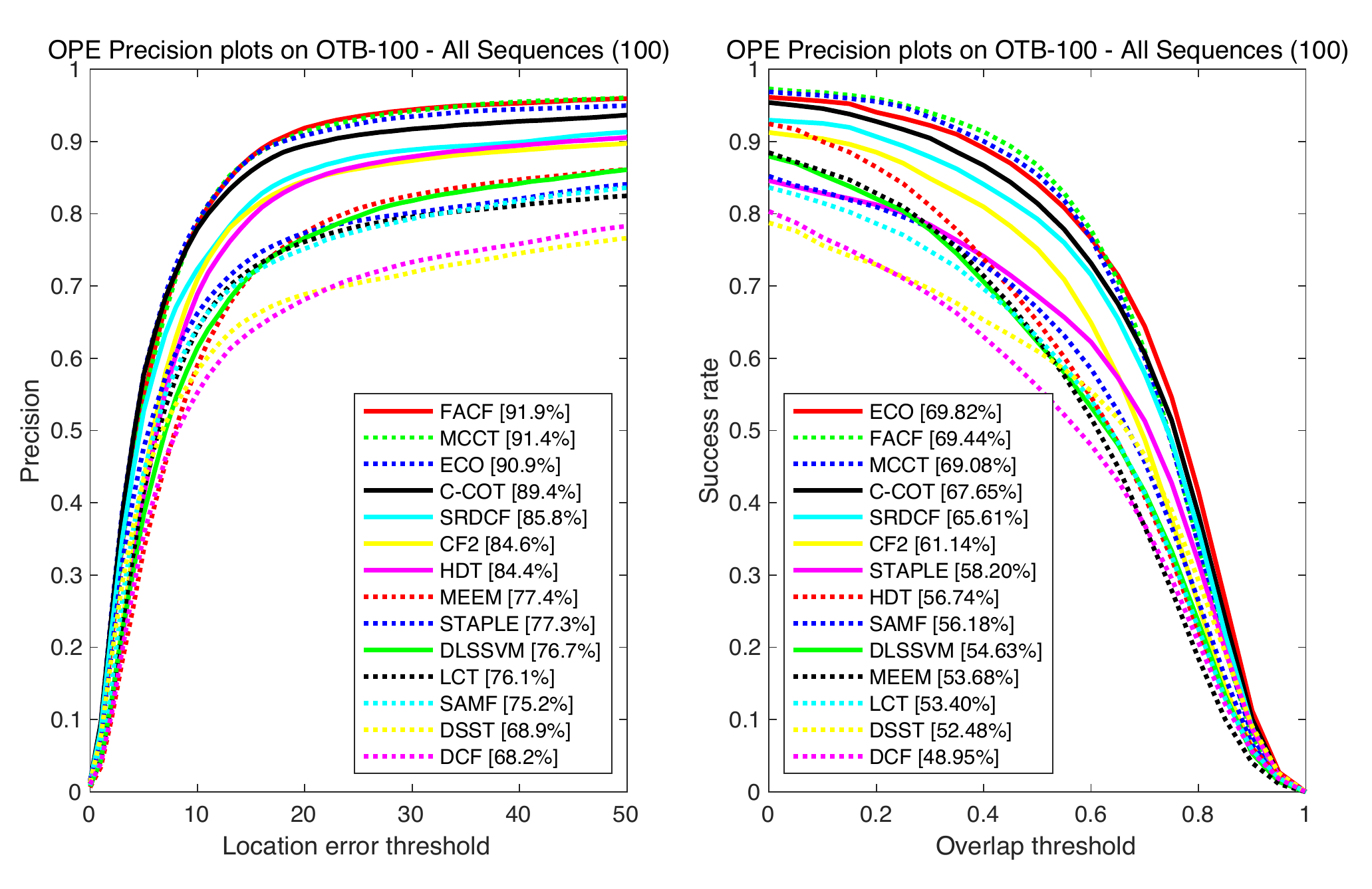}
\end{center}
   \caption{Precision and success plots on OTB-2015 \cite{2015OTB100}.  }
\label{fig:cvpr2015}
\end{figure}

\begin{table*}[t]
\caption{Performance comparison on EAO, Acc. and Rob. of the prosed method (FACF) and the state-of-the-art methods on VOT2017. }
\renewcommand\tabcolsep{3.7pt}
\renewcommand\arraystretch{1.2}
\begin{center}
\begin{tabular}{|l|c|c|c|c|c|c|c|c|c|}
\hline
~              &\tabincell{c}{DSST \cite{DSST}}& \tabincell{c}{STAPLE \cite{CVPR2016STAPLE}}  & \tabincell{c}{SiamFC \cite{SiamFC}} & \tabincell{c}{MEEM \cite{ECCV2014MEEM}} & \tabincell{c}{ECOhc \cite{CVPR2017ECO}}  &  \tabincell{c}{C-COT \cite{ECCV2016C-COT}} &  \tabincell{c}{MCCT \cite{CVPR2018MCCT}} & \tabincell{c}{\textbf{FACF} (Ours)} \\
\hline
\hline
EAO   &0.079& 0.169     &  0.188           &0.193           & 0.238    & 0.267    &0.270  &\textbf{0.274}\\
Acc.   &0.390& \textbf{0.524}    & 0.500             &0.456        & 0.489    & 0.485    &0.523  &0.494\\
Rob.    &1.494& 0.688    &0.585               &0.534         &0.435   &  0.318  &0.323    &\textbf{0.310}\\
\hline
\end{tabular}
\end{center}
\label{tab:last}
\end{table*}

To put the tracking performance into perspective, we compare the proposed best performed FACF tracker with recent state-of-the-art trackers, including DSST \cite{DSST}, MEEM \cite{ECCV2014MEEM}, SAMF \cite{ECCV2014SAMF}, KCF \cite{TPAMI2014KCF}, LCT \cite{CVPR2015LCT}, HCF \cite{HCF}, SRDCF \cite{SRDCF}, HDT \cite{Hedged}, STAPLE\cite{CVPR2016STAPLE}, C-COT \cite{ECCV2016C-COT}, ECO \cite{CVPR2017ECO}, MCCT \cite{CVPR2018MCCT}, DLSSVM \cite{CVPR2016DLSSVM} on three public benchmarks TempleColor \cite{TC}, OTB-2015 \cite{2013OTB50} and VOT2017 \cite{VOT2017}. 



\subsubsection{Evaluation on TempleColor}The precision and success plots of compared methods on the TempleColor benchmark are exhibited in Fig. \ref{fig:tc}. The proposed FACF tracker achieves the best results on P(20px) and AUC with scores of $82.2\%$ and $60.97\%$, respectively. It outperforms the second best method by $2.2\%$ and $1.99\%$. From Fig. \ref{fig:tc}, the proposed FACF significantly outperforms the recently proposed MCCT, ECO and C-COT on metrics of P(20px) and AUC and achieves the state-of-the-art performance.


\subsubsection{Evaluation on OTB-2015}
The precision and success plots of compared methods on the OTB-2015 benchmark are exhibited in Fig. \ref{fig:cvpr2015}. On this benchmark, our proposed FACF tracker achieves the best P(20px) of $93.2\%$. Among all trackers, only ECO exhibits excellent results and performs slightly better than ours on AUC metric, which is mainly due to the effectiveness of unique sample selection strategy. Our FACF tracker performs much better than C-COT, HCF, HDT, MCCT, and MEEM, which are also based on correlation filters. 

\subsubsection{Evaluation on VOT2017}The results on VOT2017 in terms of EAO, Acc., and Rob. are listed in Table \ref{tab:last}. According to EAO and Rob., the proposed FACF achieves the best performance among the recent state-of-the-art trackers. But according to Acc., STAPLE achieves the best performance. However, FACF significantly outperforms STAPLE on EAO. and Rob. which demonstrates the state-of-the-art performance of the proposed method.

\section{Conclusions}
\label{conclusion}

In this paper, we propose a hierarchical feature-aware tracking framework (FACF) based on CF for robust visual tracking, which considers not only feature-level fusion but also decision-level fusion and feature-adaptive selection to fully explore the strength of multiple features. Our framework maintains multiple experts to track the target via different views and selects the reliable outputs to refine the tracking results. Moreover, the proposed method evaluates the unreliable samples through considering the divergence of multiple experts and updates them adaptively. Through extensive experiments on several challenging data sets, we show that after adopting our simple yet effective framework, FACF is able to perform favorably against state-of-the-art ensembled methods in both accuracy and efficiency. The future work include but is not limited to attempting to deal with more complex scenarios.


\bibliographystyle{IEEEtran}
\bibliography{references}














\begin{IEEEbiography}[{\includegraphics[width=1in,height=1.25in,clip,keepaspectratio]{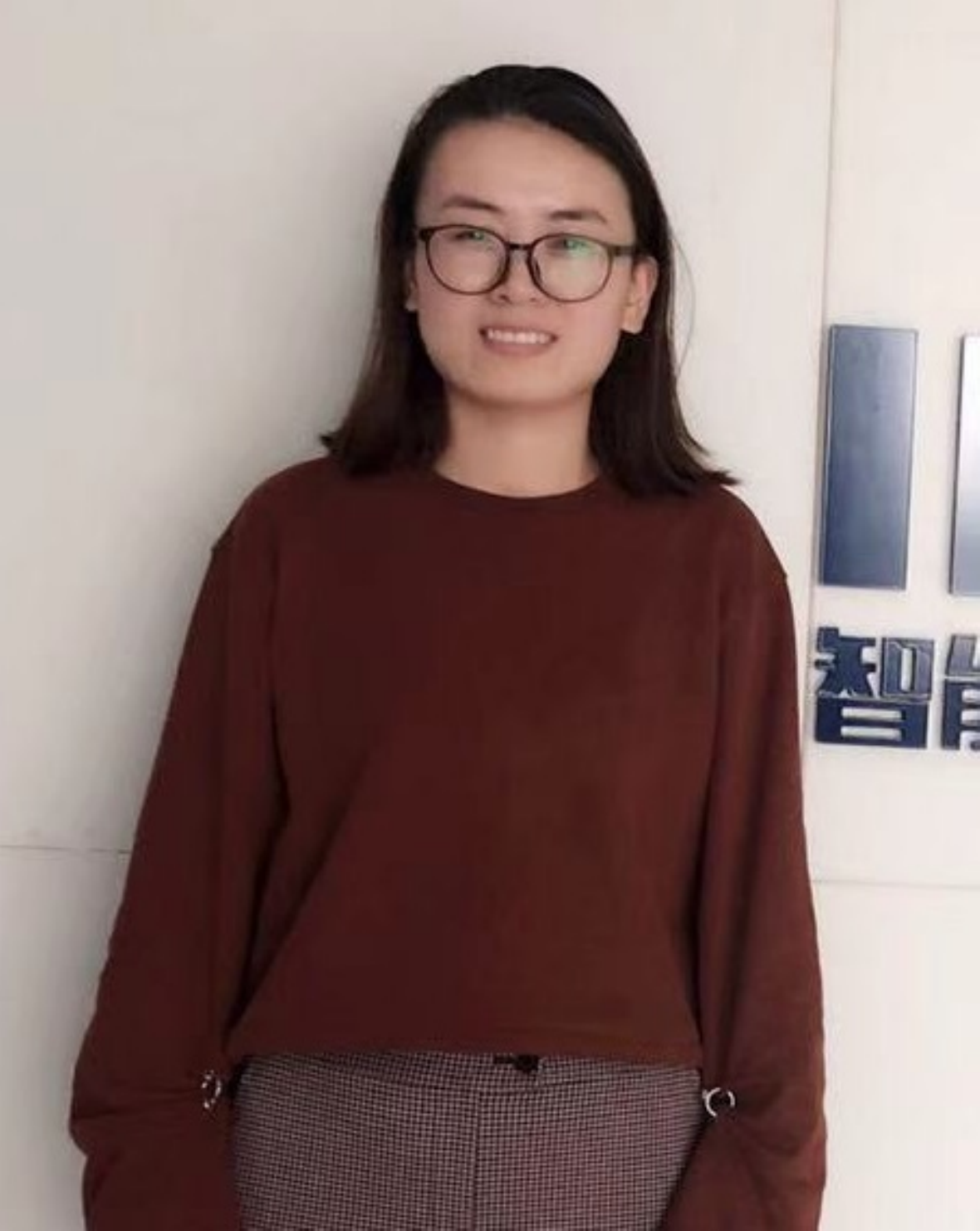}}]{Wenhua Zhang}
received the B.S. degree in communication engineering from North University of China, Taiyuan, China, in 2015. She is currently pursuing the Ph.D. degree in the School of Artificial Intelligence, Xidian Universaity, Xi'an, China.
Her research interests include machine learning and image processing.
\end{IEEEbiography}

\begin{IEEEbiography}[{\includegraphics[width=1in,height=1.25in,clip,keepaspectratio]{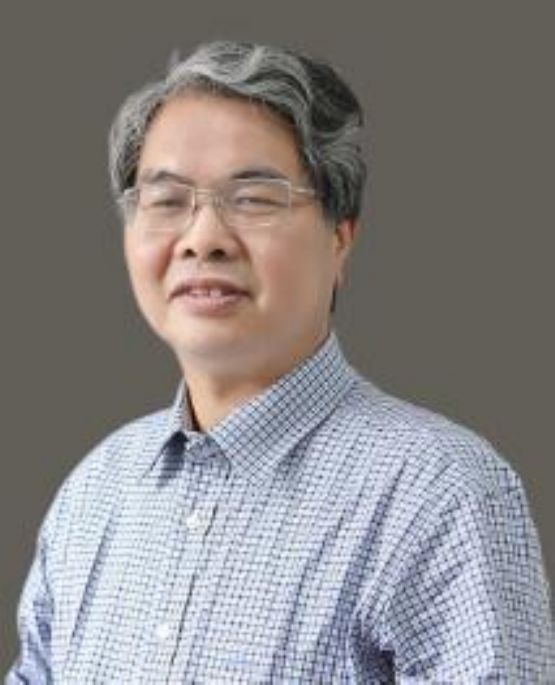}}]{Licheng Jiao}
(F'2017) received the B.S.degree from Shanghai Jiaotong University, Shanghai, China,
in 1982 and the M.S. and PhD degree from Xi'an Jiaotong University, Xi'an, China,
in 1984 and 1990, respectively.
Since 1992, he has been a Professor with the School of Artificial Intelligence,
Xidian University, Xi'an, where he is currently the Director of Key Laboratory
of Intelligent Perception and Image Understanding of the Ministry of Education
of China. His research interests include image processing, natural computation,
machine learning, and intelligent information processing.
Dr. Jiao is a member of the IEEE Xi'an Section Execution Committee, the Chairman
of the Awards and Recognition Committee, the Vice Board Chairperson of the Chinese
Association of Artificial Intelligence, a Councilor of the Chinese Institute of Electronics,
a committee member of the Chinese Committee of Neural Networks, and an expert of the
Academic Degrees Committee of the State Council.
\end{IEEEbiography}

\begin{IEEEbiography}
[{\includegraphics[width=1in,height=1.25in,clip,keepaspectratio]{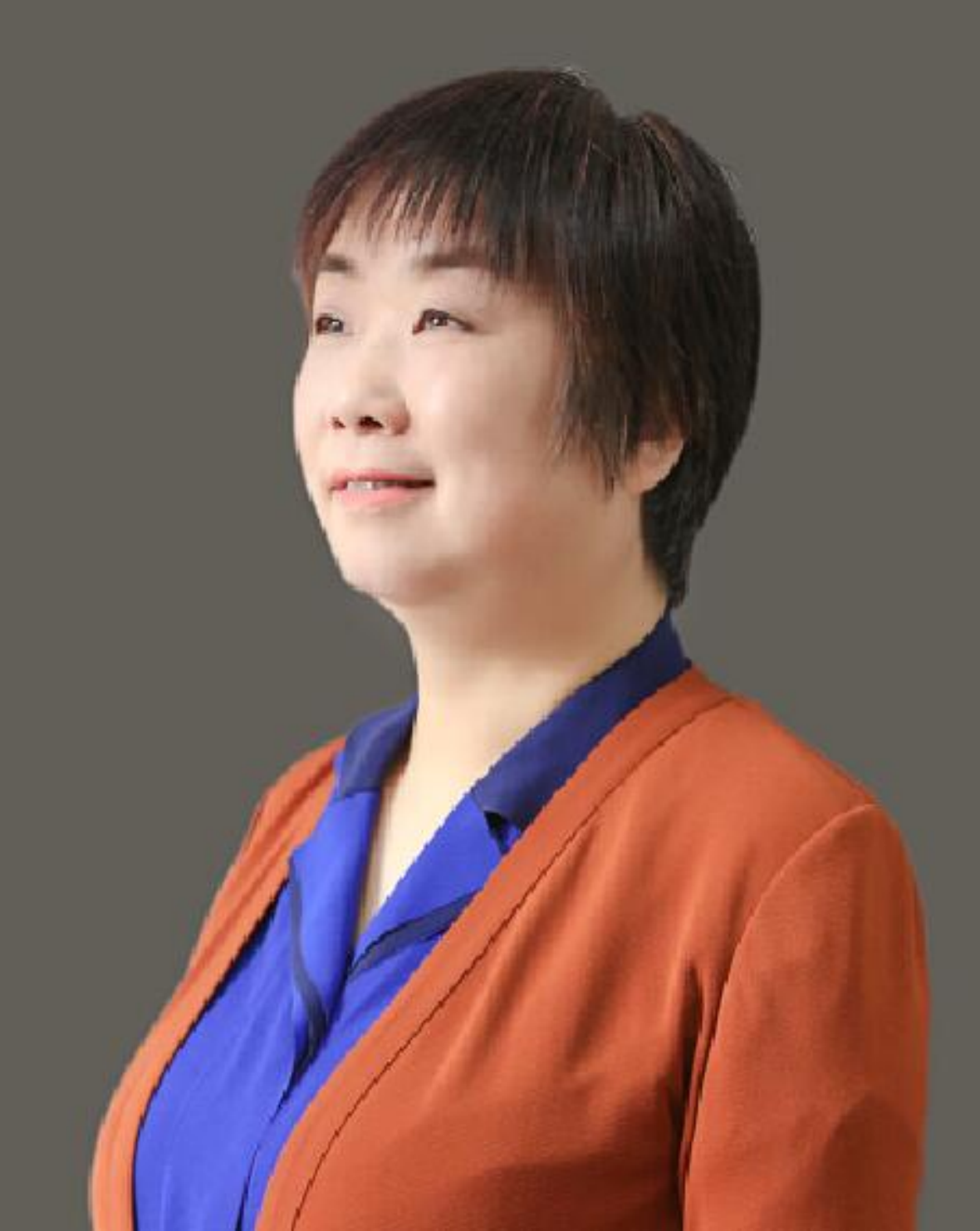}}]{Fang Liu}
(SM'07) received the B.S. degree in computer science and technology from Xi'an
Jiaotong University, Xi'an, China, in 1984 and the M.S. degree in computer science and technology
from Xidian University, Xi'an, in 1995. She is currently a Professor with the School
of Computer Science, Xidian University. Her research interests include signal and image processing,
synthetic aperture radar image processing, multiscale geometry analysis, learning theory and algorithms,
optimization problems, and data mining.
\end{IEEEbiography}

\begin{IEEEbiography}[{\includegraphics[width=1in,height=1.25in,clip,keepaspectratio]{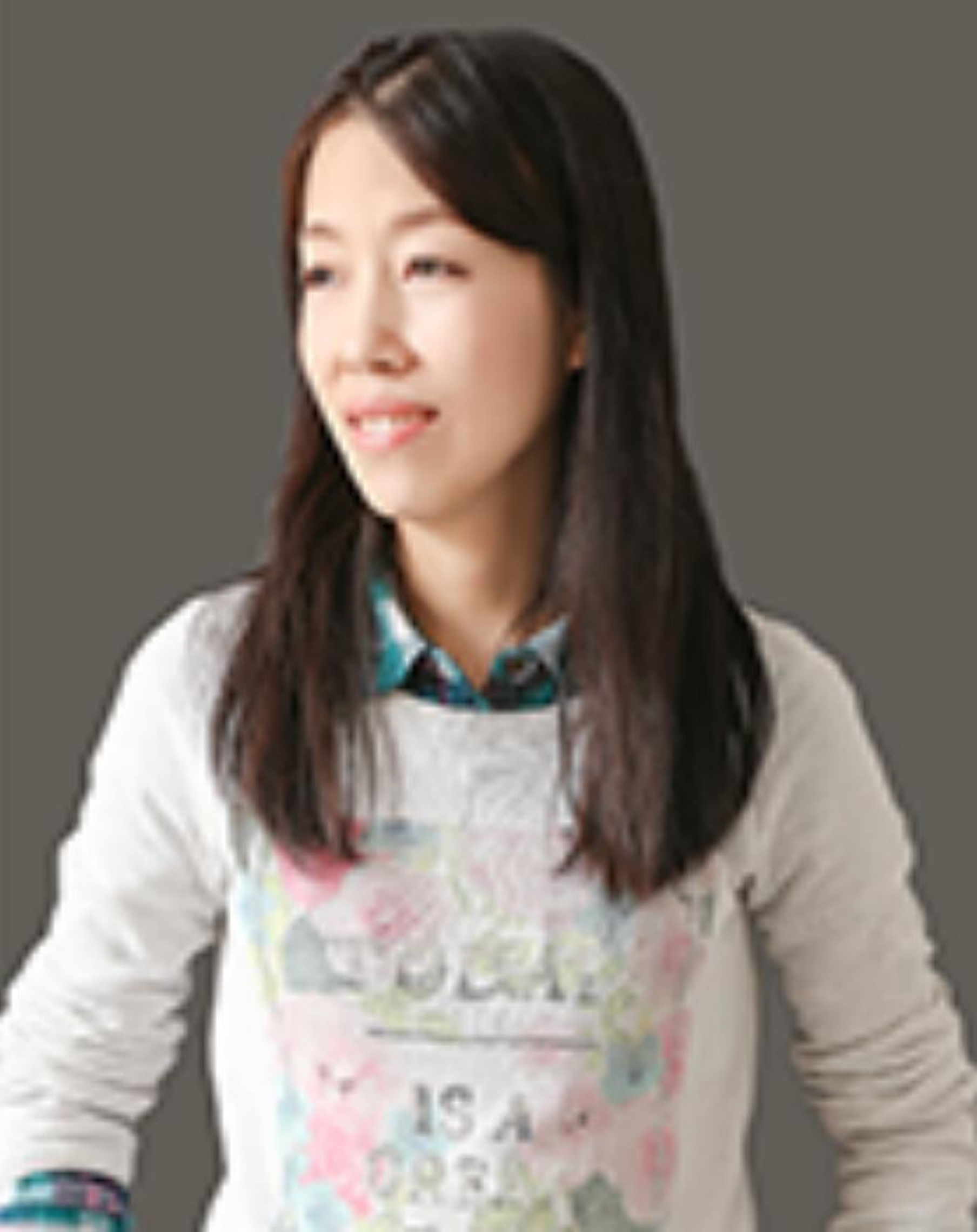}}]
{Shuyuan Yang}
(SM'14) received the B.A. degree in electrical engineering, the M.S. and Ph.D. degrees in circuit and system from Xidian University, Xi’an, China, in 2000, 2003, and 2005, respectively.
She has been a Professor with the School of Artificial Intelligence,
Xidian University, Xi'an, China. Her research interests include machine learning and multiscale geometric analysis.
 \end{IEEEbiography}

\begin{IEEEbiography}[{\includegraphics[width=1in,height=1.25in,clip,keepaspectratio]{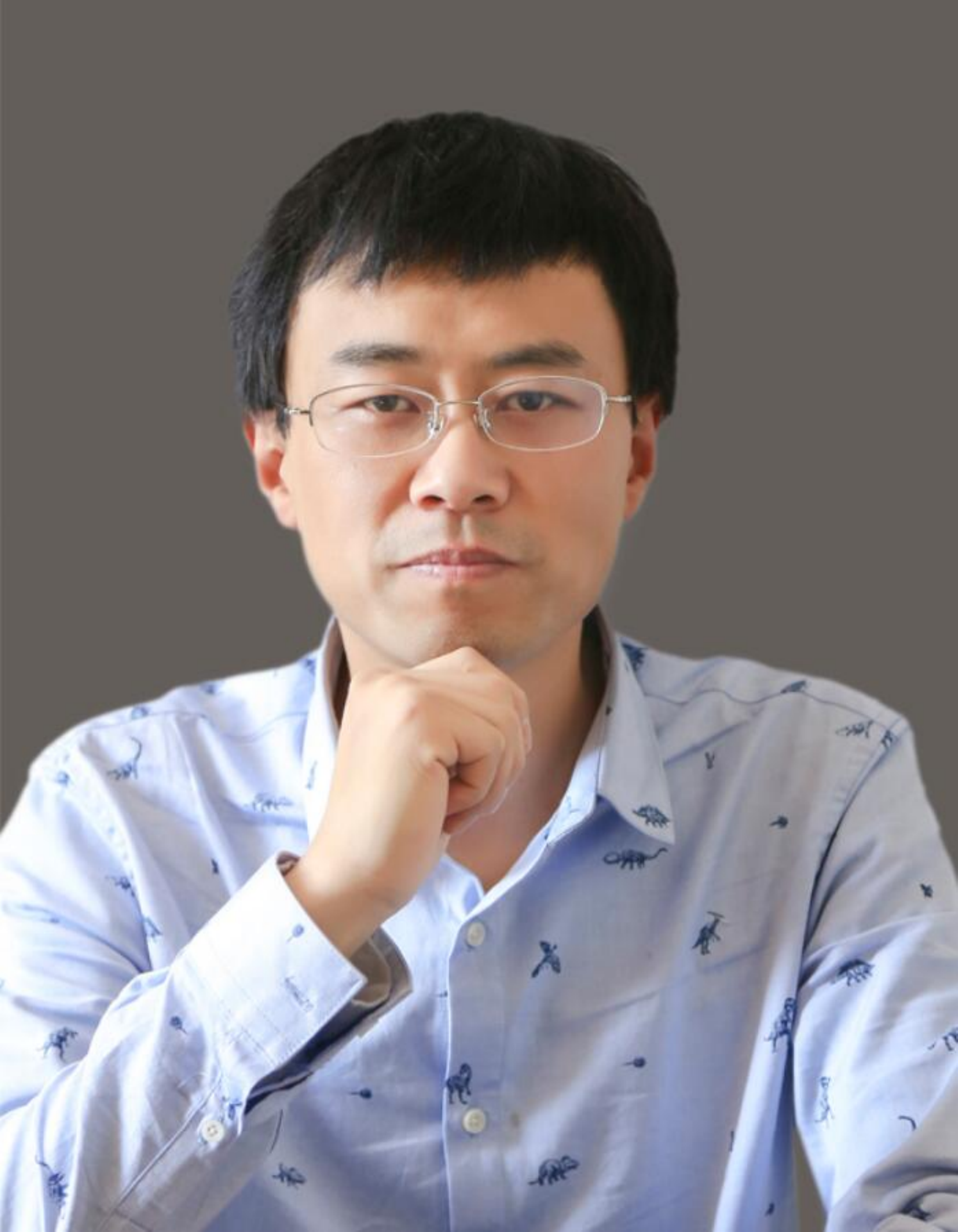}}]
{Biao Hou}
(M'07) received the B.S. and M.S. degrees in mathematics from Northwest University, Xi’an, China, in 1996 and 1999, respectively, and the Ph.D. degree in circuits and systems from Xidian University, Xi’an, China, in 2003.
Since 2003, he has been with the Key Laboratory of Intelligent Perception and Image Understanding of the Ministry of Education, Xidian University, where he is currently a Professor. His research interests include multiscale geometric analysis and synthetic aperture radar image processing.
 \end{IEEEbiography}


\end{document}